\documentclass{article}

\usepackage{hyperref}

\usepackage{PRIMEarxiv}

\usepackage{amsmath,amsfonts}
\usepackage{array}
\usepackage[caption=false,font=normalsize,labelfont=sf,textfont=sf]{subfig}
\usepackage{textcomp}
\usepackage{stfloats}
\usepackage{url}
\usepackage{verbatim}
\usepackage{graphicx}

\usepackage{color}

\usepackage{amssymb}
\usepackage{nomencl}
\usepackage{amsmath}
\usepackage{amsthm}
\usepackage{dsfont}
\usepackage{nicematrix}
\usepackage{algorithm}
\usepackage{algpseudocode}
\usepackage{tabularx, ragged2e}
\usepackage{mathtools}
\usepackage{graphicx}
\usepackage{amsfonts}
\usepackage{rotating}
\usepackage{multirow}
\usepackage{arydshln}
\usepackage{stmaryrd}
\usepackage{pgfplots}  
\usepackage{adjustbox}
\usepackage{rotating}
\usepackage{lscape}
\usepackage{afterpage}
\usepackage{pifont}

\pgfplotsset{compat=newest}

\newtheorem{proposition}{Proposition}
\newtheorem{definition}{Definition}
\definecolor{greenG}{RGB}{0,128,0}

\definecolor{lightblue}{RGB}{30,144,255}
\definecolor{aliceblue}{rgb}{0.97, 0.99, 1.0}
\definecolor{dodgerblue}{RGB}{30,144,255}
\definecolor{lightgray}{rgb}{0.9, 0.9, 0.9}
\definecolor{color1}{RGB}{30,144,255}
\definecolor{color2}{RGB}{255, 128, 0}
\definecolor{color3}{RGB}{50, 205, 50}
\definecolor{color1}{RGB}{196, 164, 132} % brown
\definecolor{color2}{RGB}{30,144,255} % blue
\definecolor{color3}{RGB}{255, 16, 240} % pink
\definecolor{greenG}{RGB}{0,128,0}
\definecolor{color_bis}{RGB}{135,206,250} % blue dark
\definecolor{color_bis2}{RGB}{0,33,243} % blue light

\title{On normalization-equivariance  properties of supervised and unsupervised denoising methods: a survey}

\author{Sébastien Herbreteau, Charles Kervrann \\
Centre Inria de l'Université de Rennes, France\\
\texttt{\{sebastien.herbreteau, charles.kervrann\}@inria.fr} \\
}

\begin{document}

\maketitle

\begin{abstract}

Image denoising is probably the oldest and still one of the most active research topic in image processing. Many methodological concepts have been introduced in the past decades and have improved performances significantly in recent years, especially with the emergence of convolutional neural networks and supervised deep learning. In this paper, we propose a survey of guided tour of supervised and unsupervised learning methods for image denoising, classifying the main principles elaborated during this evolution, with a particular concern given to recent developments in supervised learning. It is conceived as a tutorial organizing in a comprehensive framework current approaches. We give insights on the rationales and limitations of the most performant methods in the literature, and we highlight the common features between many of them. Finally, we focus on on the normalization equivariance properties that is surprisingly not guaranteed with most of supervised methods. It is of paramount importance that intensity shifting or scaling applied to the input image  results in a corresponding change in the denoiser output. 
 
\end{abstract} 

%\begin{IEEEkeywords}
%image denoising, convolutional neural networks, patch-based methods, risk minimization, equivariance.
%\end{IEEEkeywords}

\section{Introduction}

%\IEEEPARstart{I}{n}

In the realm of digital image acquisition, two significant independent types of noise can degrade the quality of captured images \cite{photon_noise, DND, unprocessing, cycleisp, poisson_estimation}. On the first hand, shot noise stems from the inherent random nature of light. Classically, the number of photons detected by an image sensor is described by a Poisson distribution whose parameter is proportional to both the true signal value and the time exposure. On the other hand, read noise is typically independent of the light intensity hitting the sensor and is introduced during the process of converting the analog signal from the camera sensor into a digital representation. Read noise is caused by various factors, including the electronic components of the sensor, circuitry, and analog-to-digital conversion process. Traditionally, this type of noise is mathematically modeled by an additive white Gaussian noise, justified by the application of the central limit theorem.

Therefore, image noise is commonly described by a mixed Poisson-Gaussian model. Formally, representing a grayscale image with $n$ pixels by a vector of $\mathbb{R}^n$ where each entry encodes the pixel intensity, the noise model is: 
\begin{equation}
    y \sim a \mathcal{P}(x/a) + \mathcal{N}(\mathbf{0}_n, b I_n)\,,
    \label{pgnoise_intro}
\end{equation}
\noindent where $y \in \mathbb{R}^n$ is the observed noisy image, $x \in \mathbb{R}^n$ is the noise-free image (true signal), and $a, b \in \mathbb{R}^+_\ast$  are the parameters relative to shot and read noise, respectively, depending in particular on the acquisition system and on the exposure time. A widespread simpler alternative to the mixed Poisson-Gaussian model (\ref{pgnoise_intro}) is the additive white Gaussian noise (AWGN) model:
\begin{equation}
    y \sim \mathcal{N}(x, \sigma^2 I_n)\,,
    \label{awgn_intro}
\end{equation}
where $\sigma^2$ is the signal-independent variance of the noise. The formulation (\ref{awgn_intro}) can be seen as an approximation of (\ref{pgnoise_intro}) where the signal-dependent shot noise is neglected. Although it may seem to be a limitation of this model, formulation  (\ref{pgnoise_intro}) actually transposes to (\ref{awgn_intro}) when using a variance-stabilizing transformation (VST) such as the Anscombe transform \cite{Anscombe}  and its generalizations \cite{Starck98,BoulangerTMI2010,MakitaloTIP2011} that amount to applying per-pixel nonlinearities that effectively reduce the signal dependence. Ultimately, due to its mathematical convenience, the AWGN model is the most widely-used one.

From the noisy observation $y$, which follows either (\ref{pgnoise_intro}) or (\ref{awgn_intro}) but also any other, possibly unknown, noise distribution, the aim of image denoising is to design a method for estimating the original unknown signal $x$ as faithfully as possible \cite{ReviewElad2023}. This amounts to identifying a function $f: \mathbb{R}^n \mapsto \mathbb{R}^n$ such that a noisy observation $y$ can be mapped to a satisfactory estimate of $x$, \textit{i.e.} $f(y) \approx x$. 
Over the years, a rich variety of strategies, tools and theories have emerged to address the issue of image denoising at the intersection of statistics, signal processing, optimization and functional analysis. 
The performance and the limitations of resulting single-shot methods are generally well understood. But this field has been recently immensely influenced by the development of machine learning techniques and artificial intelligence. Viewing denoising as a simple regression problem, this task ultimately amounts to learn to match the corrupted image to its source. The very best methods in image denoising leverage deep neural networks which are trained on large external datasets consisting of clean/noisy image pairs (see Fig.~\ref{PSNRcpu}).
However, though fast and efficient, these supervised networks suffer from their lack of interpretability and usually have fewer good mathematical properties than their conventional counterparts. 
Therefore, it is of paramount importance to examine several mathematical properties which are desirable in image denoising, especially the so-called normalization-equivariance, which ensures that any change of the input noisy image, whether by shifting or scaling, results in a corresponding change in the denoising response. 
While this property is partially fulfilled by single-shot methods, current deep neural networks surprisingly do not guarantee such a property, which can be detrimental in many situations (source of misinterpretation in critical applications). 

The remainder of the paper is organized as follows. In Section II, we take the reader on a guided tour of supervised learning methods for image denoising. In Section III, we review the unsupervised denoising methods and focus on the most performant methods. In Section IV, we study the normalization equivariance (NE) properties of the reviewed methods and provide cues so that NE holds by design.

\pgfdeclareplotmark{starBlue}{
 \color{blue}
    \pgfpathcircle{\pgfpointorigin}{3pt}\pgfusepath{fill}
    \color{white}
    \pgfpathcircle{\pgfpointorigin}{2pt}\pgfusepath{fill}
}
%\pgfdeclareplotmark{starRed}{
%    \node[draw=red,solid,fill=red] {};
%}

\pgfdeclareplotmark{starRed}{
    \color{red}
    \pgfpathcircle{\pgfpointorigin}{3pt}\pgfusepath{fill}
    \color{white}
    \pgfpathcircle{\pgfpointorigin}{2pt}\pgfusepath{fill}
}

%\addtolength{\tabcolsep}{-6pt} 

\begin{figure}[t]
\centering
\begin{tikzpicture}[scale=0.8]
\begin{axis}[
    title style={align=center},
    title={},
    cycle list name=exotic,
    ticks=both,
    %xmode=log,
    ymode=log,
    %log ticks with fixed point,
    ymin = 0.6,
    xmin = 29.04,
    xmax = 30.4,
    %y coord trafo/.code={\pgfmathparse{sqrt(#1)}},
    %y coord inv trafo/.code={\pgfmathparse{#1*#1}},
    %x coord trafo/.code={\pgfmathparse{sqrt(#1)}},
    %x coord inv trafo/.code={\pgfmathparse{#1*#1}},
    %ytick={8, 12, 16, 20, 24, 28, 32},
    xtick={29, 29.2, 29.4, 29.6, 29.8, 30, 30.2, 30.4},
    axis x line = bottom,
    axis y line = left,
    axis line style={-|},
    %nodes near coords = \rotatebox{45}{{\pgfmathprintnumber[fixed zerofill, precision=1]{\pgfplotspointmeta}}},
    nodes near coords align={vertical},
    every node near coord/.append style={font=\tiny, xshift=-0.5mm},
    ylabel={Execution time (in seconds)},
    xlabel={Average PSNR on Set12 and BSD68 (dB)},
    %xtick=data,
    %ymajorgrids, % for grids in gray
    %xmajorgrids,
    legend style={at={(1, 1)}, anchor=north east, legend columns=1},
    every axis legend/.append style={nodes={right}, inner sep = 0.2cm},
   %x tick label style={align=center, yshift=-0.6cm},
    %enlarge x limits=0.08,
    enlarge y limits=0.09,
    enlarge x limits=0.05,
    width=14cm,
    height=7cm,
]

    \addplot[only marks, blue,mark=starBlue, mark size=3pt ,mark options={solid}] coordinates {(29.335, 7.39)} node[midway, above] {\small \textcolor{black}{NL-Ridge}};

    \addplot[only marks, red,mark=starRed, mark size=3pt ,mark options={solid}] coordinates {(30.30, 52.39)} node[midway, left] {\small \textcolor{black}{Restormer}};

    \addplot[only marks, blue,mark=starBlue, mark size=3pt ,mark options={solid}] coordinates {(29.555, 54.63)} node[midway, left] {\small {\textcolor{black}{LIChI}}};
    
    \addplot[only marks, blue,mark=starBlue, mark size=3pt ,mark options={solid}] coordinates {(29.29, 0.911)} node[midway, left] {\small \textcolor{black}{NL-Bayes}};

    %\addplot[only marks, blue,mark=triangle*, mark size=3pt ,mark options={solid}] coordinates {(1, 1.96)} node[midway, right] {\small \textcolor{black}{DCT $16\times16$}};

    \addplot[only marks, blue,mark=starBlue, mark size=3pt ,mark options={solid}] coordinates {(29.27, 5.39)} node[midway, left] {\small \textcolor{black}{BM3D}};

    \addplot[only marks, blue,mark=starBlue, mark size=3pt ,mark options={solid}] coordinates {(29.545, 443.22)} node[midway, left] {\small \textcolor{black}{WNNM}};

    \addplot[only marks, red,mark=starRed, mark size=3pt ,mark options={solid}] coordinates {(30.21, 12.48)} node[midway, left] {\small \textcolor{black}{DRUNet}};

    %\addplot[only marks, blue,mark=triangle*, mark size=3pt ,mark options={solid}] coordinates {(28561, 1.56)} node[midway, left] {\small \textcolor{black}{DCT2net}};

    %\addplot[only marks, blue,mark=triangle*, mark size=3pt ,mark options={solid}] coordinates {(467233, 0.835)} node[midway, left] {\small \textcolor{black}{CARE}};

    \addplot[only marks, red,mark=starRed, mark size=3pt ,mark options={solid}] coordinates {(29.835, 3.59)} node[midway, left] {\small \textcolor{black}{DnCNN}};

    \addplot[only marks, blue,mark=starBlue, mark size=3pt ,mark options={solid}] coordinates {(29.31, 14400)} node[midway, left] {\small \textcolor{black}{S2S}};

    %\addplot[only marks, blue,mark=triangle*, mark size=3pt ,mark options={solid}] coordinates {(27.065, 1200)} node[midway, right] {\small \textcolor{black}{DIP}};

    %\addplot[only marks, blue,mark=triangle*, mark size=3pt ,mark options={solid}] coordinates {(28.18, 1200)} node[midway, right] {\small \textcolor{black}{N2S}};

    \addplot[only marks, blue,mark=starBlue, mark size=3pt ,mark options={solid}] coordinates {(29.285, 1645)} node[midway, left] {\small \textcolor{black}{RDIP}};

    \addplot[only marks, red,mark=starRed, mark size=3pt ,mark options={solid}] coordinates {(29.81, 0.77)} node[midway, left] {\small \textcolor{black}{FFDNet}};

    \addplot[only marks, red,mark=starRed, mark size=3pt ,mark options={solid}] coordinates {(29.76, 155.86)} node[midway, left] {\small \textcolor{black}{LIDIA}};

     \addplot[only marks, red,mark=starRed, mark size=3pt ,mark options={solid}] coordinates {(30.32, 14.43)} node[midway, above] {\small \textcolor{black}{SCUNet}};

          %\addplot[only marks, red,mark=starRed, mark size=3pt ,mark options={solid}] coordinates {(29.925, 14.43)} node[midway, above] {\small \textcolor{black}{N$^3$Net}};

    \addplot[only marks, blue,mark=starBlue, mark size=3pt ,mark options={solid}] coordinates {(29.32, 160.93)} node[midway, left] {\small \textcolor{black}{SS-GMM} };

    \addplot[only marks, blue,mark=starBlue, mark size=3pt ,mark options={solid}] coordinates {(29.48, 199.30)} node[midway, left] {\small \textcolor{black}{TWSC}};

    %\addplot[dashed,color=black]coordinates {(29.72,0.01)(29.72,1000) };

    \addplot[dashed,color=gray]coordinates {(0,1000)(29.58,1000)};

    \addplot[dashed,
    color=gray]
    coordinates {
    (29.58,0.01)(29.58,1000)
    };

    \addplot[only marks,color=white,mark=*, mark size=3pt ,mark options={solid}] coordinates {(29.45, 5000)} node[midway, right] { \textcolor{black}{\textit{Neural network-based}}};

    \addplot[only marks,color=white,mark=*, mark size=3pt ,mark options={solid}] coordinates {(29.48, 3.2)} node[midway, right] {\begin{sideways} \textcolor{black}{\textit{Traditional}} \end{sideways}};

\legend{\small Single-image, \small Dataset-based}

\end{axis}
\end{tikzpicture}

\label{PSNRcpu}
\caption{Execution time on CPU for images of size $512\times512$ v.s the average PSNR results on the union of Set12 and BSD68 datasets for Gaussian noise with $\sigma=25$ for  popular methods.}

\end{figure}
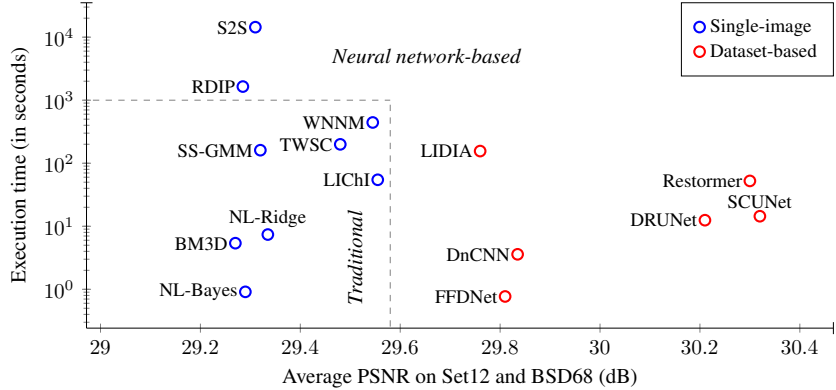

%\footnotemark  \,
%\footnotetext{\cite{berkeley}}

\section{Review of supervised learning methods}
Starting from a general framework based on empirical risk minimization, we present the three main classes of parameterized functions, also known as neural network architectures in artificial intelligence. For each architecture, we study a popular state-of-the-art representative for image denoising. Next, we address the issue of finding the best function for denoising among a given family of parametric functions, more commonly known as parameter training. Finally, we study the special case of weakly supervised learning, which does not require noise-free images for training

%------------------------------------------------------------------------------------------------------------------------------------
% SECTION 1
%------------------------------------------------------------------------------------------------------------------------------------

\subsection{Principle of supervised learning}
\label{principle_of_supervised_learning}

The holy grail in image denoising is to find a universal function $f$ that, given a noisy observation $y \in \mathcal{Y}$, maps the corresponding noise-free image $x \in \mathcal{X}$. Unfortunately, such a function is purely hypothetical as image denoising is an ill-posed inverse problem in the sense that the mere experimental observation of a noisy image is not enough to perfectly determine the unknown true image. In order to narrow down the space of possibilities and arrive at a unique solution, a risk minimization point of view has been widely adopted in past years. More precisely, let us define the risk of function $f$ as:
\begin{equation}
    \mathcal{R}(f) = \mathbb{E}_{x, y} \| f(y) - x\|\,,
    \label{f_risk}
\end{equation}
\noindent where $(x, y) \in \mathcal{X} \times \mathcal{Y}$ model all possible pairs of clean/noisy \textit{natural} images, with the associated joint probability distribution $p(x, y)$. One wants ideally to find:
\begin{equation}
    f^\ast \in \arg \min_{f} \mathcal{R}(f).
    \label{f_opt}
\end{equation}

\noindent Usually the squared $\ell_2$ norm or the $\ell_1$ norm are used to measure closeness in (\ref{f_risk}) and are examples of so-called loss functions. In the case of the squared $\ell_2$ norm, $f^\ast(y)$ is nothing else than the minimum mean square error (MMSE) estimator. Restricting $f$ to be a member of a sufficiently general class of parameterized functions $(f_\theta)$, the problem (\ref{f_opt}) transposes to the following parameter optimization problem:
\begin{equation}
    \theta^\ast \in \arg \min_{\theta} \mathcal{R}(f_\theta)\,.
    \label{theta_opt}
\end{equation}

In general, as the joint distribution $p(x, y)$ is unknown, an empirical sample consisting of a finite number $S$ of pairs of clean/noisy images, called \textit{training set}, is used as a surrogate. The empirical risk is then defined as:
 \begin{equation}
    \mathcal{R}_{\text{emp}}(f_\theta) = \frac{1}{S} \sum_{s=1}^{S} \| f_\theta(y_s)- x_s \|\,.
    \label{f_risk_emp}
\end{equation}

\noindent Note that, depending on the standard chosen to measure proximity, minimizing the empirical risk (\ref{f_risk_emp}) with respect to $\theta$ actually amounts to minimizing the mean square error (MSE) or the mean absolute error (MAE), in most cases, over a finite set of image pairs. This approach is said to be supervised in the sense that it relies on an external dataset of clean/noisy pairs of images on which the model is optimized. However, minimizing the risk on a finite subset of $\mathcal{X} \times \mathcal{Y}$, designating all possible pairs of clean/noisy images, cannot guarantee that the model will provide also good performance on unseen samples. Indeed, a function $f_\theta$ that presents a low empirical risk (\ref{f_risk_emp}) may sometimes be far from optimality with regard to the true risk defined in (\ref{f_risk}). This well-known phenomenon is called overfitting and may basically occur either when the \textit{training set} is not enough representative of the true distribution $p(x,y)$ of data in $\mathcal{X} \times \mathcal{Y}$, or when $(f_\theta)$ is over-parameterized such that it may match too closely or even exactly the \textit{training set} (in this latter case, we say that the function interpolates the data points). In that respect, optimization needs to be differentiated from machine learning which is precisely concerned with minimizing the loss on samples outside the \textit{training set}. 

Machine learning theory states that a necessary condition for good generalization beyond the \textit{training set}, is that this latter must provide sufficiently diverse, abundant and representative examples of $\mathcal{X} \times \mathcal{Y}$. Collecting high-quality \textit{training sets} may be very challenging in some situations, but the success of supervised learning depends on it, and image denoising is no exception \cite{unprocessing, cycleisp, PNGAN, scunet, fully_synthetic_training}. In order to assess the generalization capabilities of the learned model, a  \textit{test set} is used, consisting of a finite subset drawn randomly from $\mathcal{X} \times \mathcal{Y}$ and strictly disjoint from the \textit{training set} on which optimization is done. The performance of the model on the \textit{test set} is  an imperfect measure of its generalization as there exists no finite subset of $\mathcal{X} \times \mathcal{Y}$ that represents perfectly the true distribution $p(x,y)$ but it is the only  metric at our disposal.  

From this very general paradigm, several issues need be addressed. First of all, the choice of the class of parameterized functions $(f_\theta)$ is an important part of the success of supervised machine learning. The chosen class must indeed be sufficiently large for a chance to contain high-performance functions for the denoising task; but at the same time, oversized classes may lead to an overfitted model. Then, once the parameterized class of functions has been chosen, solving the inherent optimization problem defined in (\ref{theta_opt}) can be particularly cumbersome and one would like to be able to rely on efficient and general heuristics to deal with it. Finally, the quality of the \textit{training set} is crucial but, in numerous contexts, sufficiently many diverse and abundant noise-free images are unfortunately not available. A recent line of research proposes to relax the need for clean images by adopting a so-called \textit{weakly} supervised learning approach. Below, we show how all these issues are commonly addressed in the case of image denoising.

%------------------------------------------------------------------------------------------------------------------------------------
% SECTION 2
%------------------------------------------------------------------------------------------------------------------------------------

\subsection{Classes of parameterized functions}

In this section, we review the most three major classes of parameterized functions $(f_\theta)$ that were successfully experimented in image denoising. All of them are in fact subcategories of the general class of parameterized functions that are called (\textit{improperly?}) ``artificial neural networks''.

\subsubsection{Multi-layer perceptron (MLP)}

Historically, the first class of parameterized functions used in supervised machine learning is the multi-layer perceptron (MLP) proposed by F. Rosenblatt \cite{mlp_inventor}. The seminal work from H. C. Burger \textit{et al.} \cite{mlp} constitutes the first successful attempt of learning the mapping from a noisy image to its corresponding noise-free one with such an artificial neural network. For the first time in the field of image denoising, learning approaches  have been compared favorably with unsupervised (\textit{a.k.a} non-learning) methods,  without making any assumptions about natural images or about noise type.

\paragraph{Mathematical description}

Formally, a multi-layer perceptron with $L \geq 1$ hidden layers is a nonlinear function $f_\theta : y \in \mathbb{R}^{n_0} \mapsto \mathbb{R}^{n_{L+1}}$ of the following form:
\begin{equation}
    f_\theta(y) \!=\! \left[ \varphi_{\Theta_{L+1}, b_{L+1}} \circ \xi_{L} \!\circ\! \varphi_{\Theta_{L}, b_{L}} \!\circ\!  \ldots \!\circ \xi_{1} \!\circ\! \varphi_{\Theta_1, b_1} \right]\!(y),
    \label{mlp_equation}
\end{equation}
\noindent composed of:
\begin{itemize}
\item  $L+1$ parameterized affine functions $\varphi_{\Theta_l, b_l}: z \in \mathbb{R}^{n_{l-1}} \mapsto \Theta_l z + b_l$, where $\Theta_l \in \mathbb{R}^{n_l \times n_{l-1}}$ and $b_l \in \mathbb{R}^{n_l}$ are the weight matrices and bias vectors, respectively, that parameterize the MLP: $\theta = \bigcup_{l=1}^{L+1} \{ \Theta_l, b_l \}$, %\begin{equation} \theta = \bigcup_{l=1}^{L+1} \{ \Theta_l, b_l \}\,,
%\end{equation}
\item  $L$ nonlinear functions $\xi_l$ that operate component-wise. 
\end{itemize}

Interestingly, any function $f_\theta$ belonging to the MLP class in (\ref{mlp_equation}) can be viewed as a neural network. Indeed, by definition, the $L$ intermediate vectors 
\begin{equation}
    h^{(l)} = \left[\xi_{l} \circ  \varphi_{\Theta_l, b_l} \circ \ldots \circ \xi_{1} \circ \varphi_{\Theta_1, b_1}\right](y) \in \mathbb{R}^{n_l}\,
\end{equation}
 \noindent are called hidden layers and their components are referred to as hidden neurons. In the same way, vectors $h^{(0)} = y$ and $h^{(L+1)} = f_\theta(y)$ are called input layer and output layer, respectively, and their components are named neurons as well, for the sake of consistency. Moreover, the components of matrices $\Theta_l$ can be viewed as neural connections since the $i^{th}$ row of $\Theta_l$ basically maps all the neurons from the layer $h^{(l)}$ to the $i^{th}$ neuron of the following layer $h^{(l+1)}$. Finally, the nonlinear functions $\xi_l$ are called \textit{activation functions} because they aim to mimic the frequency of action potentials, or ``firing'', of real biological neurons.

\iffalse
\noindent The two historically common activation functions are the hyperbolic tangent that was used in \cite{mlp}:
\begin{equation}
    \operatorname{tanh} : t \mapsto \frac{e^t - e^{-t}}{e^t + e^{-t}} = \frac{1 - e^{-2t}}{1 + e^{-2t}}\,,
    \label{tanh}
\end{equation}
\noindent ranging from $-1$ to $1$ and the logistic sigmoid function,  ranging from $0$ to $1$, which can be derived from the previous one:
\begin{equation}
\operatorname{sigmoid} : t \mapsto \frac{1}{1 + e^{-t}} = \frac{1}{2} + \frac{1}{2} \operatorname{tanh}(\frac{t}{2})\,.
\end{equation}
\fi

Historically, the first activation functions that were investigated are the sigmoid functions, characterized by their ``S''-shaped curves. In particular the hyperbolic tangent:
\begin{equation}
    \operatorname{tanh} : t \mapsto \frac{e^t - e^{-t}}{e^t + e^{-t}} = \frac{1 - e^{-2t}}{1 + e^{-2t}}\,,
    \label{tanh}
\end{equation}
\noindent ranging from $-1$ to $1$, and its variants such as the standard logistic function were favored because they are mathematically convenient (easily computable and differentiable as $\operatorname{tanh}'(t) = 1 - \operatorname{tanh}^2(t)$) and are close to linear near origin while saturating rather quickly when getting away from it. In recent developments of deep learning the rectified linear unit (ReLU) is more frequently used as a cost-efficient alternative:
\begin{equation}
\operatorname{ReLU} : t \mapsto \max(0, t)\,.
\label{relu}
\end{equation}

\iffalse
More recently, the GELU activation function (Gaussian Error Linear Unit) \cite{gelu} was proposed as a smoother version of the ReLU with relative performance improvements in practice. It is defined as: 
\begin{equation}
\operatorname{GELU} : t \mapsto t \Phi(t)\,,
\end{equation}
\noindent where $\Phi$ is the cumulative distribution function of the standard normal distribution. Contrary to ReLU function, GELU can output negative values. Note that a useful approximation for fast computation proposed by the authors \cite{gelu} reads $\operatorname{GELU}(t) \approx t \operatorname{sigmoid}(1.702t)$.
\fi

A particularly important result \cite{mlp_universal1, mlp_universal2, mlp_universal3} states that any continuous function $f : \mathbb{R}^{n} \mapsto \mathbb{R}^{m}$ can be approximated to any given accuracy by a MLP on any compact subspace of $\mathbb{R}^{n}$, provided that sufficiently many neurons are available.  This result and its derivatives were subsequently named ``universal approximation theorems''. They all imply that neural networks can represent a wide variety of interesting functions when given appropriate weights. On the other hand, they typically do not provide a construction for the weights, but merely state that such a construction is possible.

\paragraph{MLP applied on patches for image denoising}

Given the strong mathematical guaranties provided by the ``universal approximation theorems'', the parameterized functions $(f_\theta)$ belonging to the MLP class are particularly suitable for approximating the ideal function $f$ minimizing the risk defined in (\ref{f_risk}). H. C. Burger \textit{et al.} \cite{mlp} were among the first to investigate the potential of such functions in the field of image denoising. They proposed to use a MLP to denoise the overlapping patches of noisy images, assuming that noise removal is a local issue in the images. This choice is supported by two technical observations. First of all, if MLPs were used on the entire image instead, they would be dependent on the image size which is unintended. Second, and not least, MLPs applied on the complete image would require an intractable number of parameters. Indeed, since a transition from one layer to the next requires a matrix $\Theta_l$ of parameters, the total number of parameters of a MLP is of the order of the square of the input size, in the case of constant width MLP. Transposed to images, this represents as many parameters as the square of the number of pixels! This large number of parameters makes its use prohibitive in most cases.

 The retained architecture is made up of $4$ hidden layers of size $2047$ each and is intended to be applied to patches of size  $17\times 17 = 289$. The resulting parameterized function is:
\begin{eqnarray}
    f_\theta^{\text{MLP}}: y \in \mathbb{R}^{289} \mapsto  \left[ \varphi_{\Theta_{L+1}, b_{L+1}} \circ \xi  \circ \ldots \!\circ \xi \!\circ\! \varphi_{\Theta_1, b_1} \right](y),
    \label{mlp_burger_equation}
\end{eqnarray}
\noindent where $L = 4$, the nonlinear activation function $\xi$ is the hyperbolic tangent (\ref{tanh}), and the dimensions of the weights and biases are $\Theta_1 \in \mathbb{R}^{2047 \times 289}$, $\Theta_l \in \mathbb{R}^{2047 \times 2047}$ for $2 \leq l \leq 4$, $\Theta_{5} \in \mathbb{R}^{289 \times 2047}$,  $b_{l} \in \mathbb{R}^{2047}$ for $l \leq 4$ and $b_{5} \in \mathbb{R}^{289}$. The total number of trainable parameters for this MLP is then:
$\operatorname{dim}(\theta) = 2 \times 2047 \times 289 + 3 \times 2047 \times 2047 +  4 \times 2047 + 289  = 13,762,270\,.$
For training, H. C. Burger \textit{et al.} \cite{mlp} used a large \textit{training set} of pairs of clean/noisy flattened patches ($362$ million training samples in their experiments of size $17 \times 17$ taken from the union of the LabelMe dataset \cite{dset_labelme}, containing approximately $150,000$ images, and the Berkeley Segmentation Dataset \cite{berkeley}  composed of 400 images) on which the empirical quadratic risk (\ref{f_risk_emp}) is minimized. At inference, a given noisy image is decomposed into its overlapping flattened patches and each patch is denoised separately with the learned MLP. The final denoised image is obtained by averaging the numerous estimates available for each pixel.

H. C. Burger \textit{et al.} \cite{mlp} achieved state-of-the-art results on homoscedastic Gaussian noise that compared favorably with BM3D \cite{BM3D}, the most cited unsupervised denoiser, at the cost of a full month of training on a GPU at the time. While promising, the resulting denoiser was  not yet competitive in terms of inference time and flexibility, as the model handled a single noise level and did not generalize well to other noise levels compared to other denoising methods (although solutions were proposed \cite{normalization_morel}). Moreover, the multiple artifacts induced  by the method as well as its lack of interpretability made it less usable in practice than its conventional counterparts.

\subsubsection{Convolutional neural networks (CNN)}
\label{CNN_section}

Convolutional neural networks is a class of parameterized functions $(f_\theta)$ that can be described essentially as a sparse version of multi-layer perceptrons dedicated to two-dimensional inputs. This architecture is widely used in all areas of image processing for its lightness compared to MLPs and its increased performance, image denoising being no exception \cite{tnrd, red30, drunet, dncnn, ffdnet, ircnn, mwcnn, ridnet}.

\paragraph{Mathematical description}

\begin{figure}[t]
    \centering
\includegraphics[width=.7\columnwidth]{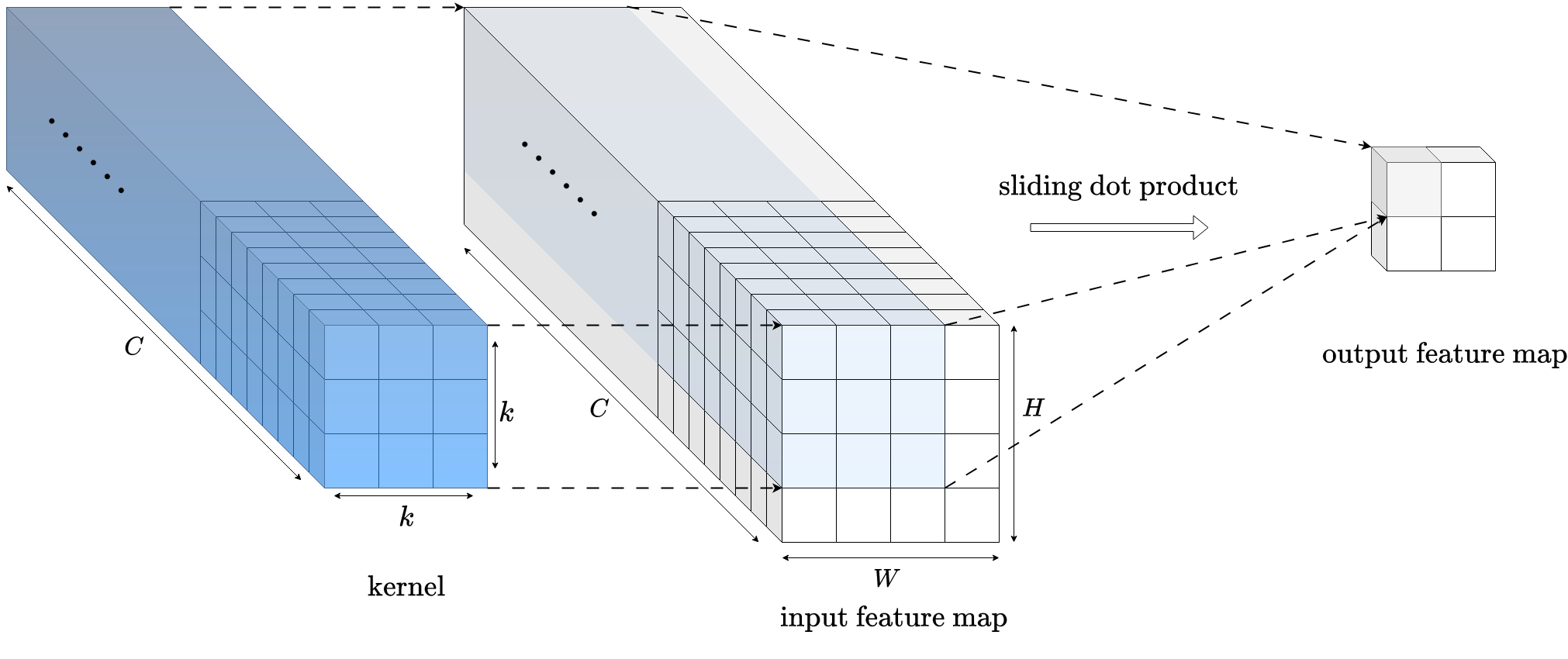}        
    \caption[Standard 2D convolution]{A $3 \times 3$ 2D convolution (without padding) producing 4 output neurons.}
    \label{figure_conv2d}
\end{figure}

The 2D convolution, or 2D cross-correlation, 
of an image $y \in \mathbb{R}^{H \times W \times C}$, or feature map, of size $H \times W$ composed of $C$ channels (color components for instance but also any abstract embedding of the input pixels) with a weight kernel $\Theta \in \mathbb{R}^{k_1 \times k_2 \times C}$ (restricted to be smaller than the dimensions of the feature map:  $k_1 \leq H$ and $k_2\leq W$), denoted $y \otimes \Theta$, is defined as a sliding dot product between $\Theta$ and the local features of $y$. This operation produces a single-channel output feature map of size $(H-k_1+1) \times (W-k_2+1)$. More precisely, a 2D convolution $y \otimes \Theta$ consists in splitting the input feature map $y$ into its overlapping 3D blocks of the same size as the kernel $\Theta$ -- there are $(H-k_1+1) \times (W-k_2+1)$ overlapping blocks -- and computing the dot product with kernel $\Theta$ for all of them: each dot product creates a pre-activated neuron. Figure \ref{figure_conv2d} illustrates the process of a 2D convolution. In practice, numerous 2D convolutions are performed successively, involving a different weight kernel $\Theta_i$ each time, and their results are concatenated along channels to produce a multi-channel output, or layer. Note that the channel size $C'$  of the output layer is strictly equal to the number of 2D convolutions that were performed. For the sake of notation simplicity, the $C'$ convolutional kernels relative to a same layer are gathered together into a unique 4D kernel, denoted by the same symbol $\Theta  \in \mathbb{R}^{k_1 \times k_2 \times C \times C'}$. Finally, a trainable vector (``bias'') $b \in \mathbb{R}^{C'}$ is generally added channel-wisely, leading to the general form of function for 2D convolutions: 
\begin{equation}
\psi_{\Theta, b}(y) = y \otimes \Theta + b\,,
\label{conv_equation}
\end{equation}
\noindent where addition applies along channels.

In some cases, it is desirable that the size $H \times W$ of the input image $y$ stays unchanged after a convolutional operation (which is generally not the case, unless $k_1=k_2=1$). A common trick to ensure size preservation is to artificially extend the size of the input image both horizontally and vertically beforehand. This operation is called padding, and the most commonly used padding strategy is simply to add zero-intensity pixels around the edges of the image: zero-padding.

Formally, a (feed-forward) convolutional neural network with $L \geq 1$ hidden layers is a nonlinear function $f_\theta: y \in \mathbb{R}^{H_0 \times W_0 \times C_0} \mapsto \mathbb{R}^{H_{L+1} \times W_{L+1} \times C_{L+1}}$ that chains 2D convolutions interspersed with nonlinear element-wise operations:
\begin{equation}
    f_\theta(y) \!=\! \left[ \psi_{\Theta_{L+1}, b_{L+1}} \!\circ\! \xi_{L} \!\circ\! \psi_{\Theta_{L}, b_{L}} \!\circ\!  \ldots \!\circ\! \xi_{1} \!\circ\! \psi_{\Theta_1, b_1} \right]\!(y),
    \label{cnn_equation}
\end{equation}
\noindent composed of:
\begin{itemize}
\item  $L+1$ parameterized convolutional  functions $\psi_{\Theta_l, b_l}: z \mapsto z \otimes  \Theta_l + b_l$, where $\Theta_l$ and $b_l$ are the weight kernels and bias, respectively, that parameterize the CNN: $\theta = \bigcup_{l=1}^{L+1} \{ \Theta_l, b_l \}$, 
\item  $L$ nonlinear functions $\xi_l$ that operate component-wise. 
\end{itemize}

Just like MLPs, the $L$ intermediate vectors:
\begin{equation}
    h^{(l)} \!=\! \left[\xi_{l} \!\circ\!  \psi_{\Theta_l, b_l} \!\circ\! \ldots \!\circ\! \xi_{1} \!\circ\! \psi_{\Theta_1, b_1}\right](y) \in \mathbb{R}^{H_l \times W_l \times C_l}\,
\end{equation}
 \noindent are called hidden layers and their components are referred to as hidden neurons. Essentially, a CNN is a MLP where affine functions are replaced with convolutional ones. A direct advantage of CNNs over MLPs is that the number of parameters is generally much smaller, as neural connections are local and identical, whatever the pixel position in the image.

Note that the basic parameterized form (\ref{cnn_equation}) of CNNs can be made more complex by adding, amongst others,  strided or dilated  convolutions \cite{dilated}, skip or residual connections \cite{resnet}, downscaling operations via pooling layers (\textit{e.g.} max pooling, average pooling...) and upscaling operations via bilinear or bicubic interpolation. A general architecture possibly incorporating all of theses features is the famous U-Net architecture \cite{unet}, widely used in computer vision.

\paragraph{Receptive field}

In a convolutional layer as shown in Fig. \ref{figure_conv2d}, each neuron receives input from only a restricted area of the previous layer called the neuron's receptive field. The receptive field has typically a 3D rectangle shape. When the network processes the input data through multiple convolutional layers, the receptive field of a neuron in deeper layers becomes larger as it incorporates information from a broader area of the input. For instance, the receptive field of a network chaining two successive $3 \times 3$ convolutional layers is the same as the receptive field of a  $5 \times 5$ convolution. The receptive field of a CNN is determined by its architectural characteristics, such as the size of the convolutional filters or the downscaling pooling operations. As moving deeper into the network, each neuron's receptive field expands due to the cascading effect of the multiple layers. Consequently, neurons in the deeper layers capture more global and complex features that encompass larger regions of the input image. Understanding the receptive field is crucial in CNNs, as it determines the spatial context that a network can capture, which is particularly essential in image denoising. Indeed, the spatial context may potentially be very useful to detect repeated patterns and denoise them properly. This is why deep CNNs with small convolutional kernels ($3 \times 3$) are widely used in computer vision; the receptive field is directly proportional to the width of the network, while the number of parameters is contained with small kernels.

\paragraph{Focus on DnCNN architecture}

\begin{figure}[t]
    \centering
\includegraphics[width=0.8\columnwidth]{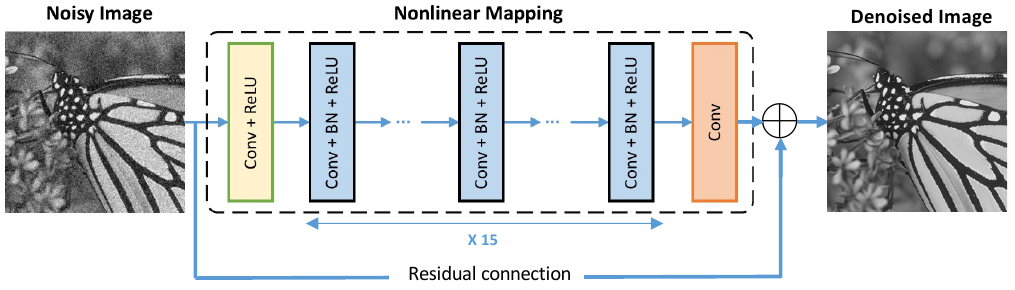}        
    \caption[The architecture of DnCNN denoising network]{The architecture of DnCNN denoising network. Source: \cite{dncnn}.}
    \label{figure_dncnn}
\end{figure}

DnCNN \cite{dncnn} (Denoising Convolutional Neural Network) is the most cited artificial neural network for image denoising so far. Its widespread popularity is due to both its simplicity and its effectiveness. Although it was developed in the early years of deep learning for image denoising, it is still considered a reference today. DnCNN is basically a feed-forward denoising convolutional neural network that chains ``conv+ReLU'' blocks, and where residual learning \cite{resnet} and batch normalization \cite{batchnorm} are utilized to speed up the training process as well as boost the denoising performance. 
Its architecture is illustrated in Figure \ref{figure_dncnn}. Formally, DnCNN encodes the following parameterized function for grayscale images:
\begin{eqnarray}
%	\begin{array}{ll}
    f_\theta^{\text{DnCNN}}:\!\!\!\!\!\!\!\!\!\!\!\!  &~& y \in \mathbb{R}^{H \times W \times 1}  \mapsto\\
	 &~& \left[ \psi_{\Theta_{L+1}, b_{L+1}} \circ \xi \circ \psi_{\Theta_{L}, b_{L}} \circ  \ldots \circ \xi \circ \psi_{\Theta_1, b_1} \right](y) + y\,, \nonumber
%	 \end{array}
    \label{dncnn_equation}
\end{eqnarray}
\noindent where $L = 16$, the nonlinear activation function $\xi$ is the ReLU function (\ref{relu}), and the dimensions of the kernels and biases are $\Theta_1 \in \mathbb{R}^{3 \times 3 \times 1 \times 64}$, $\Theta_l \in \mathbb{R}^{3 \times 3 \times 64 \times 64}$ for $2 \leq l \leq 16$, $\Theta_{17} \in \mathbb{R}^{3 \times 3 \times 64 \times 1}$,  $b_{l} \in \mathbb{R}^{64}$ for $l \leq 16$ and $b_{17} \in \mathbb{R}$. Note that the width of the hidden layers (number of channels) is arbitrarily set to $64$ for each and neither spatial upscaling, nor downscaling is used (zero-padding is leveraged all along the layers to preserve the spatial input size $H \times W$). The total number of trainable parameters for DnCNN is then: $\operatorname{dim}(\theta) = 3 \times 3 \times 64 \times 64 \times 15 + 3 \times 3 \times 64  \times 2 + 64 \times 16 + 1 = 555,137\,,$ making it much lighter than the MLP proposed by H. C. Burger \textit{et al.} \cite{mlp}. For training, the authors \cite{dncnn} used the 400 clean images from the Berkeley Segmentation Dataset \cite{berkeley} (BSD400) that they corrupted artificially with additive white Gaussian noise (AWGN) to create pairs of clean/noisy images on which the MSE is minimized. Unlike existing denoising models, which typically trained a specific set of parameters for AWGN for each noise level, DnCNN is also able, at the cost of a relatively small drop in terms of performance, to handle Gaussian denoising with an unknown noise level using a single set of parameters. This characteristic is generally referred to as ``blind'' Gaussian denoising, since the network has no knowledge of the input noise level. Moreover, the authors showed that this architecture is actually much more versatile, and can be efficiently used beyond Gaussian denoising to tackle several other inverse problems close to Gaussian image denoising. In particular, they trained a single model for three general tasks at once, namely blind Gaussian denoising, single image super-resolution (SISR) and JPEG image deblocking. For SISR, a high-resolution image is generated by first applying the bicubic upscaling on the low resolution image and then treating the inherent remaining ``error noise'' with DnCNN. Likewise, the unavoidable JPEG artifacts produced by a JPEG encoder during lossy compression are viewed as a particular type of additive noise and treated as such with the general model. Note that treating JPEG deblocking with a denoiser dedicated to Gaussian noise was already studied in \cite{jpeg_deblocking}. 

\paragraph{Focus on DRUNet architecture}

\begin{figure}[t]
    \centering
\includegraphics[width=\columnwidth]{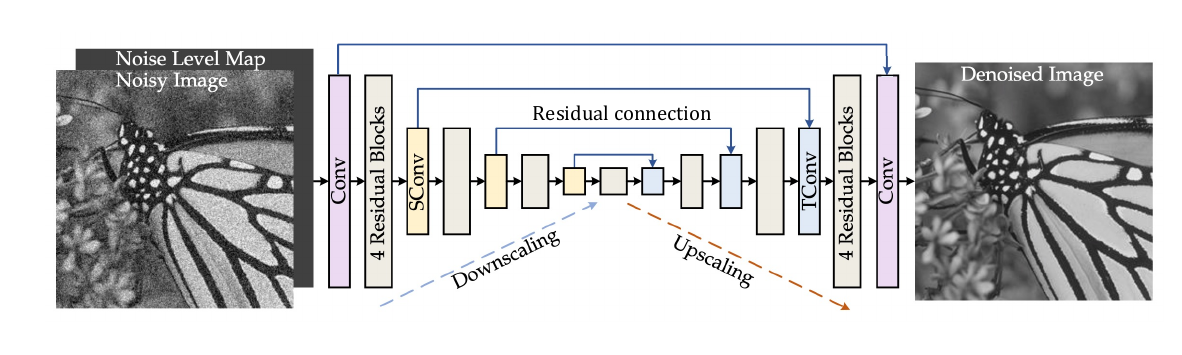}        
    \caption[The architecture of DRUNet denoising network]{The architecture of DRUNet denoising network. It takes an additional noise level map as input and combines both U-Net \cite{unet} and ResNet \cite{resnet}. ``SConv'' and ``TConv'' represent $2 \times 2$ strided convolution and transposed convolution, respectively. Source: \cite{drunet}.}
    \label{figure_drunet}
\end{figure}

More recently, DRUNet \cite{drunet} (Denoising Residual U-Net) is an architecture that was proposed as an even more competitive alternative to DnCNN \cite{dncnn}, at the price of an increased number of parameters and a longer training on a larger dataset. It achieves state-of-the-art performances for Gaussian noise removal. Contrary to DnCNN, DRUNet adopts a U-Net architecture \cite{unet}, and as such has an encoder-decoder type pathway, with residual connections \cite{resnet} all along the network. Spatial downscaling is performed using $2 \times 2$ convolutions with stride $2$ (``SConv''), while spatial upscaling leverages $2 \times 2$ transposed convolutions with stride $2$ (``TConv'') (which is equivalent to a $1 \times 1$ sub-pixel convolution \cite{pixelshuffle}). The number of channels in each layer from the first scale to the fourth scale are $64$, $128$, $256$ and $512$, respectively and each scale is composed of 4 successive residual blocks ``$3\times3$ conv + ReLU + $3\times3$ conv''. In total, the retained architecture presents $32,638,656$ parameters, which is approximately 60 times more than the number of parameters of DnCNN \cite{dncnn}, but thanks to the spatial downscaling operations, the computational complexity is contained. DRUNet architecture is illustrated in Figure \ref{figure_drunet}. Contrary to DnCNN, DRUNet is a ``non-blind'' denoiser and thus achieve increased performance over `blind'' models \cite{dncnn, ircnn}, by passing an additional noisemap as input. In the case of additive white Gaussian noise of variance $\sigma^2$, the noisemap is constant equal to $\sigma$. Note that this feature was first proposed by FFDNet \cite{ffdnet}, which is more or less the flexible ``non-blind'' variant of DnCNN \cite{dncnn}. 

Training plays a major role in the success of DRUNet. Indeed, it is widely acknowledged that convolutional neural networks generally benefit from the availability of large training data.
Therefore, the training dataset BSD400 \cite{berkeley} has been considerably enriched with the addition of many high-definition images, namely $4,744$ images from the Waterloo Exploration
Database \cite{dset_waterloo}, $900$ images from the DIV2K dataset \cite{dset_div2k}, and $2,750$ images from the Flick2K dataset \cite{dset_flickr2k}. Moreover, the authors recommend to train it by minimizing the $\ell_1$ loss instead of the mean squared error (MSE), supposedly due to its outlier robustness properties. DRUNet was trained to deal with  images corrupted with noise levels up to $\sigma=50$.

\subsubsection{Transformers}

Originally stemming from the field of natural language processing (NLP), where their introduction have led to significant improvements over convolutional neural networks, transformer-based models \cite{attention} have recently been investigated in image denoising \cite{n3net, nlrn, rnan, swinir, restormer, scunet, transformer_eccv}. This type of artificial neural network is based on the mechanism of self-attention, which allows a model to decide how important each part of an input sequence is, making it possible to find complex correlations  in the data.

\paragraph{Mathematical description}

From a multi-channel input $Y \in \mathbb{R}^{n \times m}$, where $n$ denotes the number of pixels, in the case of image denoising, and $m$ denotes the channel-size (color components for instance but also any abstract embedding of input pixels), a self-attention module produces at first three different embeddings of $Y$: queries $Q \in \mathbb{R}^{n \times l}$, keys $K \in \mathbb{R}^{n \times l}$, and values $V \in \mathbb{R}^{n \times k}$.
Traditionally, matrices $Q$, $K$ and $V$ are learned via three projection matrices $\Theta_Q \in \mathbb{R}^{m \times l}$, $\Theta_K \in \mathbb{R}^{m \times l}$ and $\Theta_V \in \mathbb{R}^{m \times k}$, such that $Q = Y \Theta_Q$, $K = Y \Theta_K$ and $V = Y \Theta_V$; but any transformation that produces the desired output shapes from $Y$ is actually suitable. Then, the self-attention is defined as:
\begin{equation}
    \operatorname{Attention}(Q, K, V) = \operatorname{softmax}(Q K^\top) V
    \label{attention_eq}
\end{equation}
\noindent where $\operatorname{softmax} : \mathbb{R}^n \mapsto \mathbb{R}^n$ is such that $\operatorname{softmax}(z)_i = e^{z_i} / \sum_{j=1}^n e^{z_j}$ and is applied over the horizontal axis in (\ref{attention_eq}). Note that $\operatorname{softmax}(Q K^\top)$ is nothing else than a right stochastic matrix of size $n$, which aims at encoding the attention weights. In others words, a self-attention module processes each entry, or ``token'', by a convex combination of all the values $V_{i, \cdot}$, weighted by the degree of attention or similarity. Moreover, it is worth noticing that the fact that $Q$  and $K$ are \textit{a priori} different matrices allows attention matrix $\operatorname{softmax}(Q K^\top)$ to be non-symmetric: token $i$ may be strongly related to token $j$ and, at the same time, token $j$ may be weakly related to token $i$ on the contrary.

The self-attention operation can actually be viewed as a general learned version of the popular NL-means \cite{nlmeans} denoiser, when rewritten as follows:
\begin{eqnarray}
	\begin{array}{ll}
    \operatorname{Attention}(Q, K, V)_{i, \cdot} = {W_i}^{-1} \sum_{j=1}^n e^{- d(Q_{i, \cdot}, K_{j, \cdot})} V_{j, \cdot} \\
	 W_i = \sum_{j=1}^n e^{- d(Q_{i, \cdot}, K_{j, \cdot})} \,, \nonumber
	 \end{array}
    \label{attention2}
\end{eqnarray}
\noindent where the \textit{pseudo} distance metric $d$ between $Q_{i, \cdot}$ and $K_{j, \cdot}$ is defined as $d(Q_{i, \cdot}, K_{j, \cdot}) = - \langle Q_{i, \cdot}, K_{j, \cdot} \rangle$. Indeed, as observed by \cite{nlrn}, the NL-Means denoiser \cite{nlmeans} is basically a transformer from the matrix of noisy patches $Y \in \mathbb{R}^{n \times m}$ ($m = p \times p$ where $p$ denotes the patch size), with identity embeddings $Q = K = Y$ and values $V = Y e_{\lceil m / 2 \rceil} = y \in \mathbb{R}^{n \times 1}$ equal to the input noisy image, and with $d$ replaced by the squared Euclidean distance: $d(Q_{i, \cdot}, K_{j, \cdot}) = \| Q_{i, \cdot} -  K_{j, \cdot} \|_2^2 / h^2$, with the hyperparameter $h$, often chosen to be proportional to the noise level $\sigma$ \cite{nlmeans2, nlmeans3, nlmeans_parameters}. As a matter of fact, even if the  distance metric $d$ was originally chosen as the opposite of the dot product between two embedded vectors for the sake of computational efficiency, the squared Euclidean distance yields comparable performance in image denoising  \cite{nlrn}.

In practice, self-attention operations cannot be applied on the entire image for the reason that the attention matrix  $\operatorname{softmax}(Q K^\top)$ in (\ref{attention_eq}) has as many entries as the squared of the input size $n$, which is in general intractable. That is why, just like MLPs (\ref{mlp_equation}), self-attention modules are deployed on subparts of the image. In general, it is used to process non-overlapping groups of neighboring embedded patches of the image \cite{swinir, restormer, scunet}. Finally, self-attention modules are usually combined with convolutional layers (\ref{conv_equation}) to get the best of both \cite{n3net, nlrn, rnan, swinir, restormer, scunet}.
 
\paragraph{Focus on SCUNet architecture}

\begin{figure}[t]
    \centering
\includegraphics[width=0.9\columnwidth]{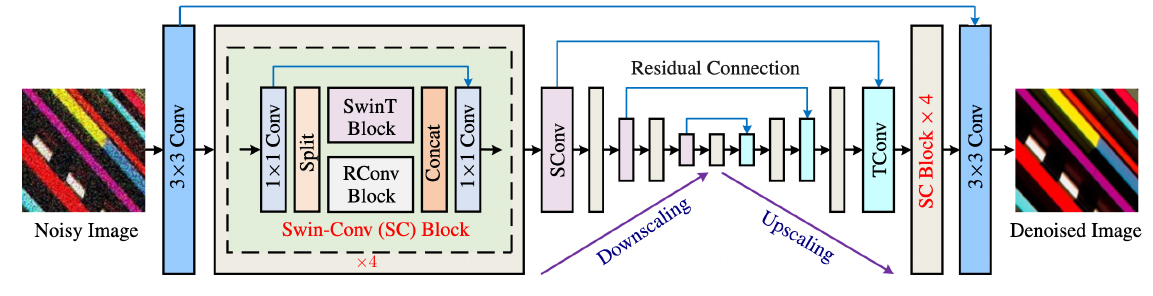}        
    \caption[The architecture of SCUNet denoising network]{The architecture of SCUNet denoising network.
    ``SConv'', ``TConv'', ``RConv'' and ``SwinT'' represent $2 \times 2$ strided convolution, $2 \times 2$ strided transposed convolution, residual ``$3 \times 3$ conv + ReLU + $3 \times 3$ conv'' block and swin transformer block, respectively. Source: \cite{scunet}.
    }
    \label{figure_scunet}
\end{figure}

Relying heavily on the DRUNet architecture (see Fig. \ref{figure_drunet}), the Swin-Conv-UNet (SCUNet) denoising network \cite{scunet} has recently been proposed as a successful attempt to incorporate self-attention modules into a convolutional neural network in order to achieve state-of-the-art performances in supervised image denoising. SCUNet basically adopts the same U-Net backbone of DRUNet and replaces the residual convolutional blocks ``$3 \times 3$ conv + ReLU + $3 \times 3$ conv'' by Swin-Conv (SC) hybrid blocks. Figure \ref{figure_scunet} summarizes the overall architecture. As illustrated, a Swin-Conv (SC) block divides in half along the channels the feature map of a $1 \times 1$ convolution to feed two independent branches, namely the ``RConv'' branch and the ``SwinT'' branch. The ``RConv'' branch is simply a residual convolutional block ``$3 \times 3$ conv + ReLU + $3 \times 3$ conv'', already used in DRUNet \cite{drunet}, with twice less parameters as in the original network, since the channel size has been halved due to the split of the feature map. As for the ``SwinT'' branch, it implements the swin transformer block described in \cite{swinir}, in turn based on the standard multi-head self-attention of the original Transformer layer \cite{attention}. Essentially, it consists in partitioning the input feature map of size $H \times W \times C$ into multiple non-overlapping groups, or windows, of equal size $(h \times w) \times c$, with $h < H$, $w < W$ and $c < C$, and processing them independently by leveraging self-attention (see formula (\ref{attention_eq})), with shared projection matrices across different windows. In the retained architecture, all windows are of equal size $(8 \times 8) \times 32$, involving self-attention matrices of size $64 \times 64$. Finally, in order to enable cross-window connections, regular and shifted window (swin) partitioning are used alternately \cite{swin}, where shifted window partitioning means shifting the feature map by $(\lfloor \frac{h}{2} \rfloor, \lfloor \frac{w}{2} \rfloor)$ pixels before partitioning. In the end, the outputs of the two branches ``RConv'' and ``SwinT'' are concatenated channel-wisely and then passed through a $1 \times 1$ convolution to produce the final residual of the input. 

Although the number of parameters of SCUNet is approximately reduced by half compared to DRUNet \cite{drunet}, since the number of parameters of ``SwinT'' blocks is negligible in relation to ``RConv'' blocks, the complexity is slightly increased, though contained. Training basically follows the instructions of DRUNet \cite{drunet}. Unlike DRUNet, SCUNet was not trained as a ''non-blind'' denoiser (\textit{i.e.} with an  additional noise level map as input), and requires instead a specific set of parameters for each noise level in the case of AWGN.

\addtolength{\tabcolsep}{-6pt} 
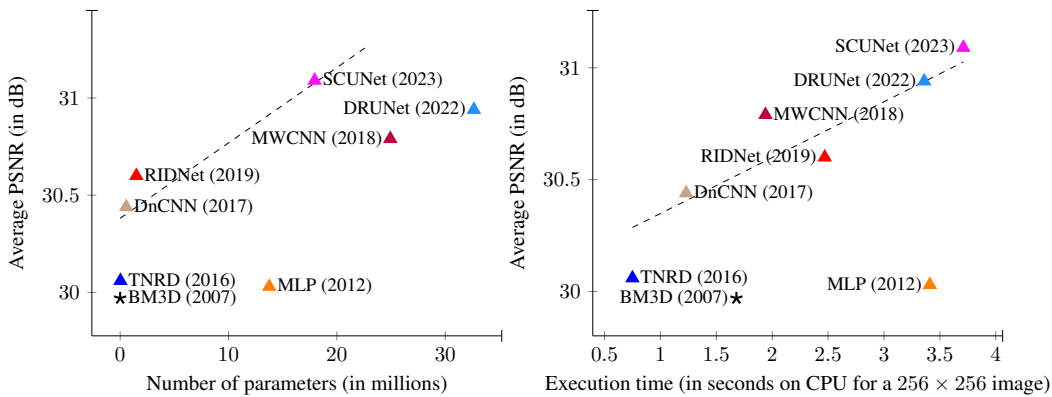
\begin{figure}[t]
\centering
\begin{tabular}{cc}
\begin{tikzpicture}[scale=0.8]
\begin{axis}[
    title style={align=center},
    title={},
    cycle list name=exotic,
    ticks=both,
    %ymin = 27.1,
    %ymax = 30.2,
    %y coord trafo/.code={\pgfmathparse{sqrt(#1)}},
    %y coord inv trafo/.code={\pgfmathparse{#1*#1}},
    %x coord trafo/.code={\pgfmathparse{sqrt(#1)}},
    %x coord inv trafo/.code={\pgfmathparse{#1*#1}},
    %ytick={8, 12, 16, 20, 24, 28, 32},
    %xtick={8, 12, 16, 20, 24, 28, 32},
    axis x line = bottom,
    axis y line = left,
    axis line style={-|},
    %nodes near coords = \rotatebox{45}{{\pgfmathprintnumber[fixed zerofill, precision=1]{\pgfplotspointmeta}}},
    nodes near coords align={vertical},
    every node near coord/.append style={font=\tiny, xshift=-0.5mm},
    ylabel={Average PSNR (in dB)},
    xlabel={Number of parameters (in millions)},
    %xtick=data,
    %ymajorgrids, % for grids in gray
    %xmajorgrids,
    legend style={at={(0.8, 0.42)}, anchor=north, legend columns=1},
    every axis legend/.append style={nodes={right}, inner sep = 0.2cm},
   %x tick label style={align=center, yshift=-0.6cm},
    enlarge x limits=0.08,
    enlarge y limits=0.15,
    width=8.4cm,
    height=7cm,
]

    \addplot[line width=1pt, orange,mark=triangle*, mark size=3pt ,mark options={solid}] coordinates {(13.762270 , 30.03)} node[midway, right] {\small \textcolor{black}{MLP (2012)}};
    
    \addplot[line width=1pt, color1,mark=triangle*, mark size=3pt ,mark options={solid}] coordinates {(0.555137, 30.44)} node[midway, right] {\small \textcolor{black}{DnCNN (2017)}};

    \addplot[line width=1pt, blue,mark=triangle*, mark size=3pt ,mark options={solid}] coordinates {(0.026885, 30.06)} node[midway, right] {\small \textcolor{black}{TNRD (2016)}};

    \addplot[line width=1pt, color2,mark=triangle*, mark size=3pt ,mark options={solid}] coordinates {(32.638656, 30.94)} node[midway, left] {\small \textcolor{black}{DRUNet (2022)}};

    \addplot[line width=1pt, red,mark=triangle*, mark size=3pt ,mark options={solid}] coordinates {(1.497041, 30.60)} node[midway, right] {\small \textcolor{black}{RIDNet (2019)}};

    \addplot[line width=1pt, purple,mark=triangle*, mark size=3pt ,mark options={solid}] coordinates {(24.927809, 30.79)} node[midway, left] {\small \textcolor{black}{MWCNN (2018)}};

    \addplot[line width=1pt, color3,mark=triangle*, mark size=3pt ,mark options={solid}] coordinates {(17.943768, 31.09)} node[midway, right] {\small \textcolor{black}{SCUNet (2023)}};

    \addplot[line width=1pt, black,mark=star, mark size=3pt ,mark options={solid}] coordinates {(0, 29.97)} node[midway, right] {\small \textcolor{black}{BM3D (2007)}};

 \addplot[domain=0:22.638656, black, dashed] {0.038772*x + 30.381295241436547};

\end{axis}
\end{tikzpicture}

& 

%~\\
 
\begin{tikzpicture}[scale=0.8]
\begin{axis}[
    title style={align=center},
    title={},
    cycle list name=exotic,
    ticks=both,
    %ymin = 27.1,
    %ymax = 30.2,
    %y coord trafo/.code={\pgfmathparse{sqrt(#1)}},
    %y coord inv trafo/.code={\pgfmathparse{#1*#1}},
    %x coord trafo/.code={\pgfmathparse{sqrt(#1)}},
    %x coord inv trafo/.code={\pgfmathparse{#1*#1}},
    %ytick={8, 12, 16, 20, 24, 28, 32},
    %xtick={1, 1.5, 2, 2.5, 3, 3.5, 4},
    axis x line = bottom,
    axis y line = left,
    axis line style={-|},
    %nodes near coords = \rotatebox{45}{{\pgfmathprintnumber[fixed zerofill, precision=1]{\pgfplotspointmeta}}},
    nodes near coords align={vertical},
    every node near coord/.append style={font=\tiny, xshift=-0.5mm},
    ylabel={Average PSNR (in dB)},
    xlabel={Execution time (in seconds on CPU for a $256 \times 256$ image)},
    %xtick=data,
    %ymajorgrids, % for grids in gray
    %xmajorgrids,
    legend style={at={(0.8, 0.42)}, anchor=north, legend columns=1},
    every axis legend/.append style={nodes={right}, inner sep = 0.2cm},
   %x tick label style={align=center, yshift=-0.6cm},
    enlarge x limits=0.12,
    enlarge y limits=0.15,
    width=8.4cm,
    height=7cm,
]

    \addplot[line width=1pt, orange,mark=triangle*, mark size=3pt ,mark options={solid}] coordinates {(3.41, 30.03)} node[midway, left] {\small \textcolor{black}{MLP (2012)}};
    
    \addplot[line width=1pt, color1,mark=triangle*, mark size=3pt ,mark options={solid}] coordinates {(1.23, 30.44)} node[midway, right] {\small \textcolor{black}{DnCNN (2017)}};

    \addplot[line width=1pt, blue,mark=triangle*, mark size=3pt ,mark options={solid}] coordinates {(0.75, 30.06)} node[midway, right] {\small \textcolor{black}{TNRD (2016)}};

    \addplot[line width=1pt, color2,mark=triangle*, mark size=3pt ,mark options={solid}] coordinates {(3.36, 30.94)} node[midway, left] {\small \textcolor{black}{DRUNet (2022)}};

    \addplot[line width=1pt, red,mark=triangle*, mark size=3pt ,mark options={solid}] coordinates {(2.47, 30.60)} node[midway, left] {\small \textcolor{black}{RIDNet (2019)}};

    \addplot[line width=1pt, purple,mark=triangle*, mark size=3pt ,mark options={solid}] coordinates {(1.94, 30.79)} node[midway, right] {\small \textcolor{black}{MWCNN (2018)}};

    \addplot[line width=1pt, color3,mark=triangle*, mark size=3pt ,mark options={solid}] coordinates {(3.71, 31.09)} node[midway, left] {\small \textcolor{black}{SCUNet (2023)}};

        \addplot[line width=1pt, black,mark=star, mark size=3pt ,mark options={solid}] coordinates {(1.68, 29.97)} node[midway, left] {\small \textcolor{black}{BM3D (2007)}};

        \addplot[domain=0.75:3.71, black, dashed] {0.24980025*x + 30.098832519986573};

\end{axis}
\end{tikzpicture}

\end{tabular}
\caption[Performance evolution of models  with the number of parameters and execution time]{Performance evolution of supervised models \cite{scunet, dncnn, ridnet, mlp, drunet, mwcnn, tnrd} with the number of parameters (left) and execution time at inference (right), respectively, for grayscale Gaussian denoising on the Set12 dataset at $\sigma=25$ (CPU: 2,3 GHz Intel Core i7). A general trend can be observed: increased performance is achieved at the cost of an increase in the number of parameters and execution time (the linear trend, in dashed line, is estimated with Theil-Sen method).}
\label{graph}
\end{figure}
\addtolength{\tabcolsep}{6pt} 

In summary, within a decade of research in supervised image denoising, the quality has been considerably enhanced, but at the price of an increased number of parameters and increased execution time (see Fig.~\ref{graph}). Nonetheless, the very best methods \cite{scunet, drunet} are now capable of recovering details barely perceptible to the human eye.

%------------------------------------------------------------------------------------------------------------------------------------
% SECTION 3
%------------------------------------------------------------------------------------------------------------------------------------

\subsection{Parameter optimization}

Once the class of parameterized functions $(f_\theta)$ -- that is the architecture of the neural network -- has been chosen, it still remains to select the best member of this class for the task of image denoising. As explained in the previous section, a proven heuristic consists in finding the optimal parameters $\theta^\ast$ that best minimize the empirical risk (\ref{f_risk_emp}), for want of knowing the true risk (\ref{f_risk}). In this section, we present the technique commonly adopted to solve this optimization problem, which is essentially based on the gradient descent algorithm.

\subsubsection{Back-propagation}

{In most of cases, the minimization of the empirical risk (\ref{f_risk_emp}) cannot be performed analytically for the reason that the chosen class of parameterized functions is in general very complex. Indeed, the resulting optimization problem (\ref{theta_opt}) is usually highly non-convex and the only fast and efficient algorithms that remain at our disposal to solve it are first-order gradient-based optimization algorithms. Calculating the gradient $\nabla_\theta \mathcal{R}_{\text{emp}}(f_\theta)$ then becomes essential.}

{The practical computation of the gradient of any weakly differentiable function at a given point $\theta$ has recently been considerably facilitated by the advent of modern machine learning libraries such as Pytorch \cite{pytorch}. Indeed, these novel frameworks are equipped with an automatic differentiation engine that powers the computation of partial derivatives. Automatic differentiation exploits the fact that the computation of a scalar value (e.g., the empirical risk \ref{f_risk_emp}) executes a sequence of elementary arithmetic operations (addition, multiplication, etc) and elementary functions (exp, square, etc). By keeping a record of data and all executed operations, partial derivatives can be computed automatically, accurately to working precision, by applying the \textit{chain rule} repeatedly to these operations. 
}

\subsubsection{Stochastic gradient descent}

{Provided with the gradient of the empirical risk with respect to the parameters $\nabla_\theta \mathcal{R}_{\text{emp}}(f_\theta)$, the most basic first-order gradient-based optimization algorithm to solve (\ref{theta_opt}) is the gradient descent algorithm. 
However, it is in practice computationally very expensive, especially for large training sets. 
An alternative method for more frequent updating is the stochastic gradient descent (SGD) \cite{sgd}. Its principle is simple: an approximation of the gradient is computed using a different random subset (mini-batch) of the entire training set at each step. With the same notations as (\ref{f_risk_emp}), $\mathcal{R}_{\text{emp}}(f_\theta)$ can be approximated by:
 \begin{equation}
\mathcal{R}^{\mathcal{B}}_{\text{emp}}(f_\theta) = \frac{1}{|\mathcal{B}|}   \sum_{s \in \mathcal{B}} \| f_\theta(y_s)- x_s \|\,,
\label{batch}
 \end{equation}
\noindent where $\mathcal{B}$ denotes a random subset of $\{1, \ldots, S\}$, so that $\nabla_\theta \mathcal{R}^{\mathcal{B}}_{\text{emp}}(f_\theta) \approx \nabla_\theta \mathcal{R}_{\text{emp}}(f_\theta)$. Then, $\nabla_\theta \mathcal{R}^{\mathcal{B}}_{\text{emp}}(f_\theta)$ can be viewed as a noisy version of the true gradient $\nabla_\theta \mathcal{R}_{\text{emp}}(f_\theta)$.  Note that, computing the gradient over a single pair of clean/noisy images $(x_s, y_s)$, can still be computationally expensive when dealing with high resolution images. This is why, $\mathcal{R}^{\mathcal{B}}_{\text{emp}}(f_\theta)$ is usually further approximated by replacing the image pairs $(x_s, y_s)$ in (\ref{batch}) by pairs of small image patches, typically of size $128 \times 128$, randomly cropped from the same images. 
The procedure is summarized in Algorithm~\ref{algo_sgd}.}

\begin{algorithm}	
\caption{Stochastic Gradient Descent (SGD) algorithm}
\small 
\begin{algorithmic}
\Require Initial parameters $\theta_0$, learning rate $\alpha$, batch size $b$, number of iterations $T$.
\Ensure Updated parameters $\theta_T$ 
\For{$t =1, \ldots, T$} 
\State Select a random subset $\mathcal{B} \subset \{1, \ldots, S\}$ of size $b$.
\State Compute gradient at point $\theta_{t-1}$: $g_t \leftarrow  \nabla_\theta \mathcal{R}^{\mathcal{B}}_{\operatorname{emp}}(f_{\theta_{t-1}})$.
\State Update parameters: $\theta_{t} \leftarrow \theta_{t-1} - \alpha  g_t$.
\EndFor 
\end{algorithmic}
\label{algo_sgd}
\end{algorithm}

\subsubsection{Adam optimization algorithm}

{Adam \cite{adam} (Adaptive Moment Estimation) is a popular extension of the stochastic gradient descent algorithm \cite{sgd}, widely used in the field of image denoising \cite{dncnn, drunet, scunet, ffdnet, n3net, nlrn} for its computational efficiency and little memory requirements. Adam combines the concepts of adaptive learning rates and momentum to provide faster convergence compared to traditional gradient descent methods, while making it less sensitive to the choice of initial learning rate. To do so, the algorithm keeps track of statistics of the first and second moment vectors, that is the gradient and its per-element square, via an exponentially decaying average. The first order moment incorporates the momentum and helps in maintaining the direction of the gradients, while the second order moment captures the magnitudes of the gradients for better adjusting the learning rates. %By correcting the bias of the first and second moment estimates to account for their initialization to zero, 
The algorithm \ref{algo_adam} provides an update rule similar to SGD \cite{sgd}. %The whole procedure is summarized in Algorithm \ref{algo_adam}. }

\begin{algorithm}
\caption{Adam algorithm}
\small 
\begin{algorithmic}
\Require Initial parameters $\theta_0$, learning rate $\alpha$, batch size $b$, number of iterations $T$, running average parameters $(\beta_1, \beta_2) = (0.9, 0.999)$, additional term for numerical stability $\varepsilon = 10^{-8}$.
\Ensure Updated parameters $\theta_T$ 
\State Initialize first and second moment vectors: $m_0 \leftarrow \mathbf{0}$ and $v_0 \leftarrow \mathbf{0}$.
\For{$t = 1, \ldots, T$} 
\State Select a random subset $\mathcal{B} \subset \{1, \ldots, S\}$ of size $b$.
\State Compute gradient at point $\theta_{t-1}$: $g_t \leftarrow  \nabla_\theta \mathcal{R}^{\mathcal{B}}_{\operatorname{emp}}(f_{\theta_{t-1}})$.
\State Update running averages: $m_t \leftarrow \beta_1 m_{t-1} + (1 - \beta_1) g_t$  and $v_t \leftarrow \beta_2 v_{t-1} + (1 - \beta_2) g_t^{\odot 2}$.
\State Compute bias-corrected moments: $\hat{m}_t \leftarrow m_{t} / (1 - \beta_1^t)$  and $\hat{v}_t \leftarrow v_{t} / (1 - \beta_2^t)$.
\State Update parameters: $\theta_{t} \leftarrow \theta_{t-1} - \alpha \hat{m}_t / (\sqrt{\hat{v}_t} + \varepsilon)$
\EndFor 
\end{algorithmic}
\label{algo_adam}
\end{algorithm}

Unfortunately, the best neural network architecture for image denoising, combined with the best optimization procedure, is powerless if high-quality clean/noisy image pairs are lacking for learning in some respects. A recent line of research tries to relax the need for clean images by adopting a so-called \textit{weakly} supervised learning approach.}

%------------------------------------------------------------------------------------------------------------------------------------
% SECTION 4
%------------------------------------------------------------------------------------------------------------------------------------

\subsection{Weakly supervised learning}

In numerous contexts, the availability of sufficiently many noise-free images is not guaranteed and supervised learning cannot be applied effectively. To circumvent this problem, attempts have been made recently to adapt empirical risk minimization (\ref{f_risk_emp}) with neural networks without ground truth. Note that, in the following, we make the arbitrary distinction between a supervised approach -- for which the \textit{training set} consists in a subset of $\mathcal{X} \times \mathcal{Y}$, designating all possible pairs of clean/noisy images, whether it is physically acquired or synthetically generated (approximated) --
 and a weakly supervised approach, for which the \textit{training set} present solely representative images from $\mathcal{Y}$.
 
\subsubsection{Learning from noisy image pairs}

A pioneer work in this spirit is Noise2Noise \cite{noise2noise} that assumes that, for the same underlying ground truth image $x_s$, two independent noisy observations $y_s$ and $\bar{y}_s$ are available. It was observed that replacing the clean/noisy pairs  $(x_s, y_s)$ by the noisy/noisy ones $(\bar{y}_s, y_s)$ in the empirical quadratic risk (\ref{f_risk_emp}) enables comparable performance to be achieved without the need for ground truths, provided that the noise is zero-mean.

A typical use case is for example fluorescence microscopy where biological cells can be fixed using a fixative agent which causes cell death, while maintaining cellular structure. By taking two successive shots of the same scene, assuming that the noise realizations are independent between them and zero-mean, it is possible to constitute a dataset composed of noisy/noisy pairs $(\bar{y}_s, y_s)$ to train a neural network $f_\theta$. Once optimized for specifically denoising  fluorescence microscopy images, the network can be deployed in a complete image processing pipeline, where noisy image pairs are no longer required (in particular, cells no longer need to be fixed).

Formally, let $f_\theta$ be a parameterized function, $x$ following the distribution of \textit{natural} images, and $y$ and $\bar{y}$ two independent random vectors following the same noise distribution from $x$ (for instance $y\sim \mathcal{N}(x, \sigma^2 I_n)$ or $y \sim \mathcal{P}(x)$). Assuming that $\mathbb{E}_{y | x}(y) =  \mathbb{E}_{\bar{y} | x}(\bar{y})= x$, we have, by developing the squared $\ell_2$ norm:
\begin{eqnarray*}
	\begin{array}{ll}
    \| f_\theta(y) \!-\! \bar{y} \|_2^2 \!\!\!\!\!&= \| (f_\theta(y) \!-\! x) - (\bar{y} \!-\! x) \|_2^2\\
	& =\| f_\theta(y) \!-\! x \|_2^2 + \| \bar{y} \!-\! x \|_2^2 \!-\! 2 \langle f_\theta(y) \!-\! x, \bar{y} \!-\! x \rangle.\!
	\end{array}
    \label{noise2noise_intro}
\end{eqnarray*}

\noindent Therefore, by taking the expected value over $x$, $y$ and $\bar{y}$:
\begin{equation}
\begin{aligned}
   \operatorname{N2N}(f_\theta) &:= \mathbb{E}_{y, \bar{y}} \| f_\theta(y) - \bar{y}\|_2^2 \\
   &\,= \mathbb{E}_{x, y} \| f_\theta(y) - x \|_2^2 + \mathbb{E}_{x, \bar{y}} \| \bar{y} - x \|_2^2\\
   & ~~~~~~~~~~~~- 2 \mathbb{E}_{x} (\mathbb{E}_{y, \bar{y} | x} \langle f_\theta(y) - x, \bar{y} - x \rangle)  \\
   &\,= \mathcal{R}(f_\theta) + \operatorname{const}  \,,
   \end{aligned}
   \label{noise2noise}
\end{equation}
\noindent  where $\mathcal{R}(f_\theta) := \mathbb{E}_{x, y} \| f_\theta(y) - x \|_2^2$ is the quadratic risk already defined in (\ref{f_risk}). Note that the expected value of the dot product cancels out since the components of $y$ and $\bar{y}$ are independent, and the noise is assumed to be zero-mean.
In the end, minimizing the risk $\mathcal{R}(f_\theta)$ amounts to minimizing the surrogate $\operatorname{N2N}(f_\theta)$ insofar as they differ by a constant value. The advantage of using $\operatorname{N2N}(f_\theta)$ is that this expression depends only on the observations $(y, \bar{y})$ and does not involve the clean images $x$ anymore. Consequently, minimizing the Noise2Noise loss is formally equivalent to minimizing the usual supervised quadratic risk. For a given neural network $f_\theta$, assuming ideal optimization, the Noise2Noise approach leads to the exact same weights $\theta^\ast$ as the supervised approach and so yields exact same performances even if it is trained without ground truth. 

However, the above reasoning assumes that an infinite amount of noisy training data is provided. In practice, for want of knowing the true risk (\ref{f_risk}), the empirical risk (\ref{f_risk_emp}) is minimized instead, and the equality (\ref{noise2noise}) does not hold for finite samples. 
Indeed, the average of dot product in (\ref{noise2noise_intro}) is as close to zero as the number of noisy data increases. Consequently, the performance of Noise2Noise drops when the amount of training data is reduced, limiting its capability in practical scenarios. In order to get the best out of Noise2Noise potential with limited noisy data, A. F. Calvarons \cite{noise2noise_improved} recently proposed to exploit the duplicity of information in the noisy pairs to generate some sort of data augmentation.

\subsubsection{Learning from single noisy images}

In certain denoising tasks, however, the acquisition of two
or more noisy copies per image can be very expensive or
impractical, in particular in medical imaging where patients
are moving during the acquisition, or in videos with moving
objects, etc. An even more remarkable line of research focuses on the possibility to train neural networks on datasets composed only of single noisy observations $y_s$.

\paragraph{SURE}
Assuming an additive white Gaussian noise model of variance $\sigma^2$, a classical result from estimation theory -- Stein's unbiased risk estimate (SURE) \cite{SURE} -- was investigated for training neural networks on datasets composed only of single noisy observations $(y_s)$ \cite{sure_loss}. Formally, let $x$ follow the distribution of \textit{natural} images and $y\sim \mathcal{N}(x, \sigma^2 I_n)$. According to \cite{SURE}, we have:
\begin{equation}
\begin{aligned}
   \operatorname{SURE}(f_\theta) \!&:=\!   \mathbb{E}_{y}   \| f_\theta(y) - y\|_2^2 + 2\sigma^2 \operatorname{div}(f_\theta)(y) - n \sigma^2 \\
   &\:=\:  \mathbb{E}_{x, y} \| f_\theta(y) - x\|_2^2   \: \: = \: \mathcal{R}(f_\theta) \,,
\end{aligned}
\label{sure_loss_}
\end{equation}
\noindent where $n$ is the dimension of images $y$ (\textit{i.e.} number of pixels). The advantage of using SURE is that the risk is expressed in such a way that it depends only on the observations $y$. Nevertheless, the SURE loss requires the computation of the divergence of $f_\theta$ at points $y$ which is cumbersome. To overcome this difficulty, the use of a fast Monte-Carlo approximation to compute the divergence term defined in \cite{monte_carlo_sure} is leveraged in \cite{sure_loss}:
\begin{equation}
\operatorname{div}(f_\theta)(y) \approx \varepsilon^\top  \frac{f_\theta(y+ h\varepsilon) - f_\theta(y)}{h} \,,
\end{equation}
\noindent where $\varepsilon$ is one single realization of the standard normal distribution $\mathcal{N}(0, I_n)$ and $h$ is a fixed small positive value.

As in the case of the N2N loss (\ref{noise2noise}),  minimizing the SURE loss is strictly equivalent to minimizing the usual supervised quadratic risk only if an infinite amount of training data is provided, which in practice does not happen. Indeed, the equality (\ref{sure_loss_}) does not hold for finite samples for similar reasons. For a sufficiently large number of data samples however, it is possible to obtain performances close to those of networks trained with ground truths.

\paragraph{Blind-spot networks}
A radical way to get rid of the divergence term is to force $f_\theta$ to be divergence-free, \textit{i.e} $\operatorname{div}(f_\theta)(y) = 0$ for all $y$. To that end, Noise2Self \cite{N2S} introduces the concept of $\mathcal{J}$-invariance. Namely, a function $f_\theta$  is said to be $\mathcal{J}$-invariant if for each subset of pixels $J \in \mathcal{J}$, the pixel values of $f_\theta(y)$ at $J$ are computed such that they do not depend on the values of $y$ at $J$. Note that such functions are in particular divergence-free
since $\frac{\partial f_\theta^{i}}{\partial y_i} (y) = 0$ for all $y$, where $f_\theta^{i}$ denotes the $i^{th}$ component of $f_\theta$. In the literature, divergence-free networks are more often referred to as blind-spot networks \cite{N2V}, as they are constrained to estimate the pixel value based on the neighboring pixels only.

Contrary to SURE loss which is limited to additive white Gaussian noise, blind-spot networks can be leveraged in a more general context. Indeed, provided that the noise is independent between pixels and is zero-mean, the minimizer the so-called self-supervised loss $\operatorname{N2S}(f_\theta) := \mathbb{E}_y \| f_\theta(y) - y\|_2^2$ is exactly the minimizer of the quadratic risk (\ref{f_risk}) \cite{N2S}. Formally, let $x$ follow the distribution of \textit{natural} images and let $y$ follow a noise distribution from $x$ which is independent between pixels (for example $y \sim \mathcal{N}(x, \sigma^2I_n)$ or $y \sim \mathcal{P}(x)$). Assuming that $\mathbb{E}_{y | x}(y) = x$, we have, by developing the squared $\ell_2$ norm:
\begin{equation*}
	\| f_\theta(y) \!-\! y \|_2^2  = \| f_\theta(y) \!-\! x \|_2^2 + \| y \!-\! x \|_2^2 - 2 \langle f_\theta(y) \!-\! x, y \!-\! x \rangle \,.
    \label{noise2self_intro}
\end{equation*}
\noindent Therefore, by taking the expected value over $x$ and $y$:
\begin{equation}
\begin{aligned}
   \operatorname{N2S}(f_\theta) &:= \mathbb{E}_{y} \| f_\theta(y) - y \|_2^2 \\
   &\,= \mathbb{E}_{x, y} \| f_\theta(y) - x \|_2^2 + \mathbb{E}_{x, y} \| y - x \|_2^2 \\
   & ~~~~~~~~~~~~- 2 \mathbb{E}_{x} (\mathbb{E}_{y | x} \langle f_\theta(y) - x, y - x \rangle)  \\
   &\,= \mathcal{R}(f_\theta) + \operatorname{const}  \,,
   \end{aligned}
   \label{noise2self}
\end{equation}
\noindent  where $\mathcal{R}(f_\theta) := \mathbb{E}_{x, y} \| f_\theta(y) - x \|_2^2$ is the quadratic risk already defined in (\ref{f_risk}). Note that the expected value of the dot product cancels out since $f_\theta$ is blind-spot, the components of $y$ are independent between pixels, and the noise is assumed to be zero-mean. Therefore, minimizing the risk $\mathcal{R}(f_\theta)$ amounts to minimizing the surrogate $\operatorname{N2S}(f_\theta)$ insofar as they differ by a constant value. An ingenious example of a divergence-free network is proposed by Noise2Kernel \cite{N2K} that exploits donut kernels for the first layer and dilated convolutional kernels for the next layers. Finally, note that Noise2Void \cite{N2V} proposed before Noise2Self \cite{N2S} the idea of using the self-supervised loss with a blind-spot network, although the theoretical justification provided was not as strong as that of \cite{N2S}.

Nevertheless, the performance of divergence-free functions is considerably limited by the constraint of not voluntarily using the information of key pixels. Indeed, except from the parts of the signal that are easily predictable (for example uniform regions), counting exclusively on the information provided by the neighborhood to denoise the pixels is an inefficient strategy. Think for example of the extreme case of a uniform black image with a single white pixel on its center. With a blind-spot network, the central white pixel will be lost and wrongly replaced by a black one.

\paragraph{Probabilistic blind-spot networks} To improve the performance of blind-spot networks, several authors \cite{pN2V, laine, convallaria} propose to refine the predictions during inference when the noise model is known. % or when it can at least be estimated. 
For this purpose, they adopt a Bayesian point of view, different from the risk minimization point of view (\ref{f_risk}), used until now. Following this paradigm, a network $f_\theta$ is trained so that, given exclusively the noisy surroundings $\Omega_y$ of a noisy pixel $y$ (the central noisy pixel $y$ is excluded), it outputs a (parameterized) probability distribution $p_\theta(x|\Omega_y)$ of the central clean pixel. In other words, $f_\theta$ is such that $f_\theta(\Omega_y)$ predicts a learned prior probability distribution of the expected central clean value, instead of just predicting a value without taking uncertainty into account, as in the risk minimization paradigm. Equipped with such a function $f_\theta$, Bayes' rule can be applied to update the prior with new information of the noisy central pixel $y$, provided that the noise model is known, to obtain the posterior  distribution:
\begin{equation}
    \underbrace{p(x| y, \Omega_y)}_{\text{posterior}} \propto \underbrace{p(y| x, \Omega_y)}_{\text{likelihood}} \underbrace{p(x|\Omega_y)}_{\text{prior}} \approx \underbrace{p(y| x)}_{\substack{\text{noise} \\ \text{model}}} \underbrace{ p_\theta(x|\Omega_y)}_{\substack{\text{learned}  \\ \text{prior}}}\,.
\end{equation}
From the posterior, the Minimum Mean Squared Error (MMSE) estimate (\textit{i.e.} the conditional expectation) or the Maximum A Posteriori (MAP) is produced, which can be considered as an improved version of the prediction given by Noise2Self \cite{N2S}, since it is refined with the information of the central pixel. Note that the adopted Bayesian point of view enables to efficiently combine the knowledge learned on an external dataset composed of noisy images and the information of the input noisy image, which would not have been possible with a risk minimization paradigm.

The remaining questions are now how to construct $f_\theta$ and how to train it. First of all, an arbitrary parametric model for the prior $p_\theta(x|\Omega_y)$ needs to be chosen. In \cite{pN2V}, $f_\theta$ is built in such a way that $f_\theta(\Omega_y)$ outputs a vector of the size of the number of different intensities of the image (a $256$-dimensional vector when images are coded on $8$ bits for example) where all entries are non-negative and sum to one, interpreted as the histogram of a discrete probability distribution. In \cite{laine}, $f_\theta(\Omega_y)$ is constrained to follow a Gaussian model and so the output simply consists in a two-dimensional vector, encoding the mean $f_\theta(\Omega_y)_1$ and standard deviation $f_\theta(\Omega_y)_2$ of a Gaussian distribution. As for training, they both use the method of Maximum Likelihood Estimation (MLE). For a data sample $\{y_s\}_{s \in \{ 1, \ldots, S \}}$ of $S$ noisy central pixels surrounded by neighborhoods $\{\Omega_{y_s}\}_{s \in \{ 1, \ldots, S\}}$, the log-likelihood function reads (using the formula of total probability):
\begin{equation}
     \ln \mathcal{L}(\theta; \{y_s\}) =    \sum_{s=1}^{S}  \ln \underbrace{\int_{-\infty}^{+\infty} p(y_s | x)  p_\theta(x | \Omega_{y_s})}_{p_\theta(y_s|\Omega_{y_s})} \; dx.
\end{equation}

\noindent In the case of an additive white Gaussian noise model of variance $\sigma^2$ and when $f_\theta(\Omega_y)$ is constrained to output a Gaussian model \cite{laine}, we have: $p(y_s | x) = \mathcal{N}(y_s; x, \sigma^2)$ and $ p_\theta(x | \Omega_{y_s}) = \mathcal{N}(x; f_\theta(\Omega_{y_s})_1, f_\theta(\Omega_{y_s})_2^2).$
It follows that 
\begin{equation}
\begin{aligned}
p_\theta(y_s|\Omega_{y_s}) &= \int_{-\infty}^{+\infty} p(y_s | x)  p_\theta(x | \Omega_{y_s}) dx \\
&= \mathcal{N}(y_s;  f_\theta(\Omega_{y_s})_1, \sigma^2 + f_\theta(\Omega_{y_s})_2^2)\,.
\end{aligned}
\end{equation}
Finally, the  solution $\theta^\ast \in \arg \max_{\theta}  \ln \mathcal{L}(\theta; \{y_s\})$ is as follows: %of the following form: 
\begin{equation}
\theta^\ast \!=\! \arg \min_{\theta} \sum_{s=1}^{S}  \ln (\sigma^2 \!+\! f_{\theta}(\Omega_{y_s})_2^2) + \frac{(y_s \!-\! f_{\theta}(\Omega_{y_s})_1)^2}{\sigma^2 \!+\! f_{\theta}(\Omega_{y_s})_2^2},
\end{equation}
which is solved using Adam algorithm \cite{adam}.

Experiments on artificially noisy images \cite{laine} but also on real-world noisy images \cite{pN2V} tend to show that weakly supervised probabilistic approaches are almost on par with their supervised counterparts.

\paragraph{Noisier2Noise}

An ingenious way of dispensing with the probabilistic approach, while making full use of the central pixel, was proposed by Noisier2Noise \cite{noisier2noise} and Recorrupted-to-Recorrupted \cite{R2R}. Their approach is based on adding more noise to single noisy images in the dataset, although this may seem  counter-intuitive. The idea of Noisier2Noise \cite{noisier2noise} is to train a network $f_\theta$ that maps the original noisy images $y$ from noisier versions $z$ synthetically generated by adding extra noise. 
The authors argue that, with this strategy, the network is encouraged to predict $\mathbb{E}(y|z)$; and $\mathbb{E}(x|z)$ can be estimated thereafter during the inference step via a linear combination of $\mathbb{E}(y|z) \approx f_{\theta^\ast}(z)$ and $z$. For example, in the most simple case  where $y = x + \varepsilon$ with $\varepsilon \sim \mathcal{N}(0, \sigma^2 I_n)$ and $z = y + \varepsilon'$ with $\varepsilon' \sim \mathcal{N}(0, \sigma^2 I_n)$ with $\varepsilon'$ independent from $\varepsilon$ , we have, by linearity of expectation and by noticing that $\mathbb{E}(\varepsilon|z) = \mathbb{E}(\varepsilon'|z)$:
\begin{equation*}
	\begin{aligned}
    2 \mathbb{E}(y|z) &= \mathbb{E}(x|z) + \left(\mathbb{E}(x|z) + \mathbb{E}(\varepsilon|z) + \mathbb{E}(\varepsilon'|z)\right)\\
	& = \mathbb{E}(x|z) + \mathbb{E}(z|z) =   \mathbb{E}(x|z) + z\,, 
	\end{aligned}
\end{equation*}
hence $\mathbb{E}(x|z) = 2 \mathbb{E}(y|z) - z$. Therefore, at inference, for a noisy observation $y$, the denoised image is finally  estimated by $2 f_{\theta^\ast}(y + \varepsilon') - (y + \varepsilon')$ where $\varepsilon'$ is a realization of $\mathcal{N}(0, \sigma^2 I_n)$.

More recently, still in the setting of additive
white Gaussian noise (AWGN) of variance $\sigma^2$, \textit{i.e.} $y \sim \mathcal{N}(x, \sigma^2 I_n)$, Recorrupted-to-recorrupted \cite{R2R} showed that it is possible, from a noisy image $y$, to construct an artificial pair of independent noisier images $(z, \bar{z})$, centered in $x$, that can be exploited to train a neural network, just like in \cite{noise2noise} (see equation (\ref{noise2noise})). In the end, a Noise2Noise-like equality holds:
\begin{equation}
  \operatorname{R2R}(f_\theta):=\mathbb{E}_{z, \bar{z}} \|  f_\theta(z) - \bar{z}\|_2^2 \; = \; \mathbb{E}_{x, z} \| f_\theta(z) - x \|_2^2 + \operatorname{const}\,, 
 \label{riskNr2N2}
\end{equation}
where $\mathbb{E}_{x, z} \| f_\theta(z) - x \|_2^2$ is a ``noisier'' risk close to the target risk $\mathcal{R}(f_\theta)$ defined in (\ref{f_risk}). Minimizing the R2R loss is then equivalent to minimizing the ``noisier'' risk. To denoise an input noisy image $y$ at inference, it is first renoised according to the recorruption model $z$ to get the final estimate $f_{\theta^\ast}(z)$. Provided that the artificial $z$ is not much noisier than $y$, this strategy achieves performances close to those of networks trained with ground truths.

Interestingly, among the different possible recorruption models, there is the straightforward setting $z = y + \alpha  \varepsilon$ and  $\bar{z} = y - \varepsilon / \alpha$, with $\varepsilon \sim \mathcal{N}(0, \sigma^2 I_n)$ and $\alpha \in \mathbb{R}^\ast$. 
According to the property of affine transformation of Gaussian vectors, we have:
\begin{eqnarray}
\begin{aligned}
\begin{pmatrix} z \\ \bar{z} \end{pmatrix} 
	 & \sim \mathcal{N}\left(\!\begin{pmatrix} x \\ x \end{pmatrix} \!,\! \begin{pmatrix}  (1+\alpha^2) \sigma^2 I_n0\! \!\!&\! \!\! \mathbf{0}_{n \times n} \\ \mathbf{0}_{n \times n}\!\! \!&\! \!\! (1+1/\alpha^2) \sigma^2 I_n \end{pmatrix}\! \right)\!,
\end{aligned}
\end{eqnarray}
meaning that $z$ and $\bar{z}$ are independent from each other. In practice, $\alpha=0.5$ is recommended for training  to balance the noise of $z$ and $\bar{z}$ \cite{R2R}.

\paragraph{Noise2Score} Finally, another original and versatile method for learning without ground truths was proposed by Noise2Score \cite{noise2score}. In this novel approach, the conditional mean of the posterior distribution $\mathbb{E}(x | y)$ (posterior expectation of $x$ given noisy observation $y$) is calculated leveraging a classical result from Bayesian
statistics, namely Tweedie’s formula \cite{tweedie}, which involves the so-called score function. Formally, assuming that the likelihood $p(y|x)$ can written under the form $p(y|x) = a(x) b(y) \exp(x^\top T(y))$ with $a: \mathbb{R}^n \mapsto \mathbb{R}$, $b: \mathbb{R}^n \mapsto \mathbb{R}$ and $T: \mathbb{R}^n \mapsto \mathbb{R}^n$ (subset of the exponential family which covers a large class of important distributions such as the Gaussian, binomial, multinomial, Poisson, gamma, and beta distributions, as well as many others), then the following equality holds:% (see proof in \textcolor{red}{Appendix \ref{tweedie_appendix}}): 
\begin{equation}
    J_T(y)^\top \mathbb{E}(x |  y) = \nabla_y \ln (p(y)) -  \nabla_y \ln (b(y)) \,,
\end{equation}
\noindent where $J_T$ denotes the Jacobian matrix of function $T$. In particular, when $T$ has the simple form $T(y) = c y$, with $c \in \mathbb{R}^\ast$, $J_T(y)^\top = c I_n$ and finally the  conditional mean of the posterior distribution is:
\begin{equation}
    \mathbb{E}(x |  y) = \left(\nabla_y \ln (p(y)) -  \nabla_y \ln (b(y))\right) / c \,,
    \label{tweedie_formula}
\end{equation}
\noindent where $\nabla_y \ln (p(y))$ is referred to as the score (gradient of the marginal distribution of $y$). 

As it stands, the formula (\ref{tweedie_formula}) is purely theoretical since the distribution of \textit{natural} noisy images $p(y)$ is at least as difficult to know as the distribution of \textit{natural} images $p(x)$. However, capitalizing on the recent finding that the score function can be stably estimated from the noisy images \cite{score_approx}, Noise2Score \cite{noise2score} suggests to use a residual denoising autoencoder $f_\theta$ for approximating the score:
\begin{equation}
	\begin{aligned}
&\nabla_y \ln (p(y)) \approx f_{\theta^\ast}(y),\\
& \theta^* \in \arg \min \; \underset{\substack{y\sim p(y),  \varepsilon \sim \mathcal{N}(0, 1) \\ \alpha \sim \mathcal{N}(0, \delta^2)}}{\mathbb{E}} \|    f_{\theta}(y+ \alpha \varepsilon) + \varepsilon/\alpha \|_2^2\,
 \end{aligned}
\label{training_noise2score}
\end{equation}
with $\delta \to 0$ (note the similarity with Recorrupted-to-recorrupted \cite{R2R} for recorrupted images $y + \alpha \varepsilon$). The advantage of Noise2Score \cite{noise2score} is that, provided that the noise model belongs to the exponential family distribution, the problem comes down to estimating the score function always approximated by the same universal training (\ref{training_noise2score}). 

\noindent %\underline{Example for AWGN:}
In the case of an additive white Gaussian noise model of variance $\sigma^2$, we have: 
\begin{equation}
p(y | x)  
= a(x) b(y) \exp(x^\top T(y))\,,
\end{equation}
\noindent with $a(x) = \left(\sigma \sqrt{2\pi}\right)^{-n} 
\exp \left(-\frac{1}{2\sigma^2}\|x\|_2^2\right)$, $b(y) = \exp \left(-\frac{1}{2\sigma^2}\|y\|_2^2\right)$ and $T(y) = y / \sigma^2$. As $\nabla_y \ln (b(y)) = - y / \sigma^2$, Tweedie's formula then reads:
\begin{equation}
\mathbb{E}(x |  y) = y + \sigma^2 \nabla_y \ln (p(y)) \approx y + \sigma^2 f_{\theta^\ast}(y) \,.
\end{equation}

\subsection{Discussion and conclusion}

In spite of their great theoretical interest, weakly supervised approaches for image denoising, which are designed to learn without ground truths, are unfortunately of limited practical value. Indeed, if collecting a dataset of noisy image pairs is assumed to be possible as in Noise2Noise \cite{noise2noise}, why not collect several $n$-tuples of noisy images instead which, once averaged, would constitute ground truth images for use in a supervised framework (approach retained for the datasets of \cite{convallaria} for example). As for learning from datasets of single noisy images, the proposed approaches are either disappointing in terms of performance \cite{N2S, N2V} due to strong architectural constraints, or, require the noise model to be known \cite{sure_loss, pN2V, laine, noisier2noise, R2R, noise2score} in order to achieve performance comparable to that of supervised models. As a consequence, weakly supervised learning is far from being the preferred strategy for tackling challenging benchmarks such as the Darmstadt Noise Dataset \cite{DND} where only single real-world noisy images are available, for which the real noise can only be roughly approximated mathematically by a mixed Poisson-Gaussian model. Instead, the best-performing methods \cite{unprocessing, cycleisp, PNGAN, scunet} simulate a large amount of realistic noisy images from clean ones by carefully considering the noise properties of image sensors, on which any denoising neural network can be trained on. The same observation can be made in fluorescence microscopy, where the most popular denoising neural  network \cite{CARE} was trained in a supervised way, whether on physically acquired or synthetic training data.

\section{Unsupervised denoising  methods}
Both supervised and weakly supervised learning strategies are extremely dependent on data quality (although they do not rely on the same type of image pairs), which is a well-established weakness. In some situations, it may be challenging to gather a large enough dataset for learning. Only unsupervised methods - in which only the noisy input image is used for training - are operationally available. Historically, these methods were studied before their supervised counterparts, partly due to the computational limitations of the time that made resource-intensive supervised learning unthinkable.  In this chapter, we present a non-exhaustive list of well-known unsupervised algorithms, classified according to four different main principles. As we shall see, the best unsupervised denoisers share key elements, in particular the property of self-similarity observed in images, whatever their category.

%------------------------------------------------------------------------------------------------------------------------------------
% SECTION 5
%------------------------------------------------------------------------------------------------------------------------------------

\subsection{Weighted averaging methods}

The most basic unsupervised methods for image denoising are without a doubt the smoothing filters, among which we can mention the averaging filter or the Gaussian filter for the linear filters and the median filter for the nonlinear ones. Interestingly, the linear smoothing filters can actually be viewed formally as elementary convolutional neural networks $f_\Theta(y) =  y \otimes \Theta$ already defined in Section II  with no bias, no hidden layer and no activation function, and with unique convolutional kernel $\Theta$.
In contrast to supervised CNNs, the kernel is non-trainable.
Note that symmetric padding is applied on the noisy image $y$ beforehand to ensure size preservation.

In practice, the smoothing filters act by replacing each intensity value of noisy pixels with a convex combination of those of its neighboring noisy pixels. Denoising is made possible, at the cost of edge blur, by reducing the variation in intensity between neighboring pixels. Although these filters are extremely rudimentary, they are sometimes used as pre-processing steps in some popular algorithms where performance is not at stake such as the Canny edge detector \cite{canny}  due to their unbeatable speed. Building on the idea of convex combinations of noisy pixels, numerous extensions were proposed by better adapting to the local structure of the images \cite{bilateral, nlmeans, OWF, yaroslavsky, averaging_filters, these}. In what follows, we review three major unsupervised denoisers \cite{bilateral, nlmeans, OWF} processing images via convex combinations of noisy pixels. 

Formally, we denote by $y$ a vectorized noisy image patch of size $n$ whose central pixel is $y_c$ (the value of index $c$ is $\left \lceil{n/2}\right \rceil$). Each method of this subsection implements a denoising function of the form $f_\theta(y) = y^\top \theta$ aimed at estimating the noise-free
central pixel $x_c$, and where the weights $\theta \in \mathbb{R}^n$ are patch-dependent and are such that $\mathbf{1}_n^\top \theta =  1$ and $\theta \succeq 0$.

\iffalse
\begin{equation}
f_{\operatorname{bilateral}}(y)_i = \frac{1}{W_i}\sum_{y_j \in \Omega(y_i)} g_r(\|y_i - y_j\|) g_s(\|i-j\|) y_j
 \quad \text{with} \quad  W_i = \sum_{y_j \in \Omega(y_i)} g_r(\|y_i - y_j\|) g_s(\|i-j\|)\,,
% \label{bilateral}
\end{equation}
\noindent where $y_i$ denotes the $i^{th}$ component of vector $y$, $\Omega(y_i)$ the set of its neighboring pixels, the abusive notation $\|i-j\|$ designates the distance between the 2D coordinates of pixels $i$ and $j$ (the Euclidean distance for example) and the non-negative real-valued functions $g_r$ and $g_s$ are the intensity range and spatial kernels, respectively. The latter functions can be Gaussian functions for example: $g_r : x \mapsto \exp(-x^2 / h_r^2)$ and $g_s : x  \mapsto \exp(-x^2 / h_s^2)$, where $h_r$ and $h_s$ are the range and spatial smoothing hyperparameters, respectively. As $h_r$ increases, $g_r$ approaches the constant function and the above function (\ref{bilateral}) has a behavior close to a Gaussian smoothing filter. On the contrary, as $h_r$ decreases, $g_r$ gives more weight to pixels with high intensity similarity and the resulting filter becomes nonlinear and more edge-preserving.
\fi

{as the bilateral filter \cite{bilateral} that evaluates the intensity similarity between two neighboring pixels, the seminal work from A. Buades \textit{et al.} \cite{nlmeans} adopts a more robust approach by exploiting the similarity of patches instead. For each pixel, an average of the neighboring noisy pixels, weighted by the degree of similarity of patches they belong to, is leveraged for edge-preserving denoising. 
The convex weights of the N(on)-L(ocal) Means can be defined as:
\begin{equation}
\theta_i = K^s_{i} K^r(\|p(y_i) - p(y_c)\|)   / \sum_{j=1}^{n}  K^s_{j} K^r(\|p(y_j) - p(y_c)\|)
\label{nlm_weights}
\end{equation} 
\noindent where $p(y_i)$ represents the vectorized patch centered at $y_i$ (whose size can be different from the size of the image patch $y$), and where
$K^s \in \mathbb{R}_+^n$ is a spatial kernel used to give more weight to pixels closer to the central pixel and  $K^r : u \mapsto \exp(-u^2 / h^2)$, where $h$ is the range smoothing hyperparameter. As $h$ increases, $K^r$ approaches the constant function and the filter has a behavior close to a Gaussian smoothing filter. On the contrary, as $h$ decreases, $K^r$ reinforces the weighting of pixels with high patch similarity and the resulting filter becomes nonlinear and more edge-preserving.
}

\iffalse
Formally, it is defined by:  
\begin{equation}
f_{\operatorname{NLM}}(y)_i = \frac{1}{W_i}\sum_{y_j \in \Omega(y_i)} e^{-\frac{\|p(y_i) - p(y_j)\|_2^2}{h^2}} y_j \quad \text{with} \quad  W_i = \sum_{y_j \in \Omega(y_i)} e^{-\frac{\|p(y_i) - p(y_j)\|_2^2}{h^2}}
\end{equation}
\noindent where $y_i$ denotes the $i^{th}$ component of vector $y$, $p(y_i)$ represents the vectorized patch centered at $y_i$, $\Omega(y_i)$ the set of its neighboring pixels and the smoothing parameter $h$ is  proportional to the noise level $\sigma$ in the case of additive white Gaussian noise as proposed by several authors \cite{nlmeans2, nlmeans3, nlmeans_parameters}. 
\fi

The resulting N(on)-L(ocal) Means \cite{nlmeans} algorithm has had a tremendous influence on the denoising field and above for the reason that it is capable of effectively process redundant information in images with the help of patches. 
The central idea is that, in a natural image, a patch rarely appears alone and that almost perfect copies can be found in its surroundings \cite{irani}. 
NL Means has paved the way for a brand new class of denoising algorithms that exploits the self-similarity assumption \cite{BM3D, bayesnlmeans_kervrann, nlbayes, WNNM, LSSC, SAIST, NCSR, PLR, nlmeans_kervrann, tupin}. 
{Nevertheless, determining the optimal weights $\theta$ of convex combinations for image denoising still remained an open question, although patch self-similarity appears to be a key element for obtaining competitive results. 
In \cite{OWF}, Jin \textit{et al.} addressed this question starting from  \cite{nlmeans,nlmeans_kervrann,localA}. They achieved state-of-the-art performances among methods restricted to convex combinations of pixels via the establishment of an upper bound for the optimal weights $\theta$. 
Adopting a risk minimization approach and constraining the weights $\theta$ to encode a convex combination of pixels, the optimal weights in the case of Gaussian noise (i.e., $y \sim \mathcal{N}(x, \sigma^2 I_n)$) and in the $\ell_2$ sense, are:
\begin{equation}
\theta^\ast = \arg \min_{\theta \in \mathbb{R}^{n}}  \mathcal{R}(f_\theta) \quad \text{s.t.} \quad  \mathbf{1}_n^\top \theta =  1  \text{ and } \theta \succeq 0 \,,
\label{owf_risk}
\end{equation}
where $\mathcal{R}(f_\theta) := \mathbb{E}_y ((f_\theta(y) - x_c )^2)$ is the quadratic risk. By leveraging a bias–variance decomposition, the statistical risk $\mathcal{R}(f_\theta) = \left( \mathbb{E}_y(f_\theta(y) - x_c) \right)^2 + \mathbb{V}_y(f_\theta(y) - x_c)$, under convex constraints, has a closed-form expression which can be upper bounded using the triangle inequality:
\begin{equation}
\begin{aligned}
    \mathcal{R}(f_\theta)  
    & \leq f_\theta(|x-x_c|)^2 + \sigma^2  \|\theta \|_2^2 
     = \theta^\top Q \theta\,,
\end{aligned}
\label{owf_upper_bound}
\end{equation}
\noindent where the subtraction applies element-wise and $Q := |x-x_c| |x-x_c|^\top + \sigma^2 I_n$ is a symmetric positive definite matrix. Finally, OWF \cite{OWF} proposes to approximate the optimal weights $\theta^\ast$ defined in (\ref{owf_risk}) by the ones minimizing the upper bound (\ref{owf_upper_bound}) under convex constraints. This amounts to solving a quadratic program and the resulting weights have a closed-form expression \cite{OWF}.
}

%------------------------------------------------------------------------------------------------------------------------------------
% SECTION 6
%------------------------------------------------------------------------------------------------------------------------------------

\subsection{Sparsity methods}

Sparsity methods have emerged as powerful tools for image denoising, offering effective ways to restore images corrupted by noise while preserving important structural information. These methods exploit the inherent sparsity of natural images, which implies that most image patches can be efficiently represented by a small number of non-zero coefficients in a suitable transform domain.

\subsubsection{Sparsity in a fixed basis}

Sparsity of patches in a fixed basis refers to the property that most image patches can be efficiently represented using only a small number of non-zero coefficients in a predetermined basis. A basis is a set of linearly independent vectors, or patches, that spans the entire signal space. It should be distinguished from the term dictionary, for which the vectors are not necessarily linearly independent.  

Formally, we denote by $x \in \mathbb{R}^n$ a vectorized clean \textit{natural} image patch of size $n$. According to the sparsity assumption, there exists a fixed basis of vectors $\{ b_i \}_{i \in \{1, \ldots, n \}}$, where $b_i \in \mathbb{R}^n$, such that each clean patch $x$ of a noise-free image can be exactly represented 
by a linear combination involving only a few basis vectors. Adopting the matrix notation where $B \in \mathbb{R}^{n \times n}$ is the matrix formed by stacking the basis vectors $\{ b_i \}_{i \in \{1, \ldots, n \}}$ along columns, the sparsity assumption reads:
\begin{equation}
    \forall x \in \mathbb{R}^n, \: x \textit{ is a natural patch} \: \Leftrightarrow  \| B^{-1}x \|_0   \leq t_0 \,,
\end{equation}
\noindent where $\| \cdot \|_0$ is the $\ell_0$ \textit{pseudo} norm counting the non-zero elements of a vector, $t_0 \leq n$ is an hyperparameter controlling the sparsity and the entries of vector $B^{-1} x$ are the unique coefficients of the linear combination which generate patch $x$ in basis $B$.

A general strategy for denoising a noisy patch $y$ under the sparsity paradigm is then to find its closest sparse representation. The resulting optimization problem is as follows:
\begin{equation}
     \arg \min_{x \in \mathbb{R}^n} \| y - x \| \quad \text{s.t.} \quad \| B^{-1} x \|_0 \leq t_0 \,,
\end{equation}
which is equivalent, thanks to the change of variable $x=B\theta$, to:
\begin{equation}
     \arg \min_{\theta \in \mathbb{R}^n} \| y - B\theta \| \quad \text{s.t.} \quad \| \theta \|_0 \leq t_0 \,.
     \label{sparsity_optimization}
\end{equation}

\noindent  Note that denoising under the sparsity assumption involves several poorly defined quantities, namely the number of non-zero coefficients $t_0$ for being considered sparse, the norm  $\| \cdot \|$ to choose for assessing the patch proximity and especially the fixed basis $B$. Common choices for the basis $B$ include the discrete cosine transform (DCT) or  wavelets \cite{DCT, wavelet, wavelet_sure} as discussed below. 

Finally, note that solving (\ref{sparsity_optimization}) exactly in the general case where $B$ is a dictionary can be done in a finite amount of computation but this is a NP-hard problem. The algorithms designed to find an approximate solution of (\ref{sparsity_optimization}) are called pursuit algorithms and include basis pursuit, FOCUSS, or matching pursuit methods \cite{matching_pursuit3, matching_pursuit2, matching_pursuit}. 

\paragraph{TV denoising} Total variation (TV) denoising \cite{TV} is finally one of the most famous image denoising algorithm, appreciated for its edge-preserving properties. In its original form \cite{TV}, a TV denoiser is defined as a function $f: \mathbb{R}^n \times \mathbb{R}^+_\ast  \mapsto \mathbb{R}^n$ that solves the following equality-constrained problem:
\begin{equation}
    f_{\operatorname{TV}}(y, \sigma) = \mathop{\arg \min}\limits_{
\substack{x \in \mathbb{R}^{n}}} \: \| x \|_{\operatorname{TV}}  \quad \text{s.t.} \quad  \| y-x \|_2^2  = n\sigma^2
\end{equation}
\noindent where $\| x \|_{\operatorname{TV}} := \| \nabla x \|_2$ is the total variation of $x \in \mathbb{R}^n$.

\paragraph{DCT and DWT denoiser}
The discrete cosine transform (DCT) algorithm \cite{DCT} is a simple and efficient sparsity-based method for image denoising. 
The DCT is closely related to the discrete Fourier transform, but involves only real numbers. This basis yields several pleasant mathematical properties; in
particular it is orthogonal, meaning that $B^{-1} = B^\top$, and there exists a fast algorithm \cite{fast_DCT}  for computing the decomposition of any vector in this basis, just like the FFT algorithm (Fast Fourier Transform). 
Secondly, the DCT basis is experimentally near optimal to approximate \textit{natural} patches in the sense that it ensures maximum energy compression of data in the first components. 
In order quickly approach the solution of the sparsity optimization problem (\ref{sparsity_optimization}), a simple procedure \cite{DCT}  consists in computing the DCT of the noisy patch, that is $B^{-1}y$, and setting to zero all small coefficients (below $3\sigma$ in absolute value for Gaussian noise). 
At the end, all denoised patches are repositioned at their initial locations and averaged to produce the final denoised image. The discrete wavelet transform (DWT) is another example of set of orthogonal bases $B$ that has been successfully utilized for image denoising  \cite{wavelet, wavelet_sure, wavelet_best, wavelet_christine}. Contrary to the DCT, these bases are itself
sparse, in the sense that a majority of coefficients of the basis vectors are zero. This property makes decomposition calculations particularly fast. Interestingly, the DWT has the
ability to decompose an image into different frequency subbands at different scales. The high-frequency subbands capture the local details and fine structures, while the low-frequency
subbands represent the global structures and smooth regions. The Bayesian denoising method BLS-GSM \cite{wavelet_best} achieved state-of-the-art performance at the time by carefully adapting noise processing to each scale.

\paragraph{BM3D} 

\begin{figure}[t]
\centering
\includegraphics[width=0.8\columnwidth]{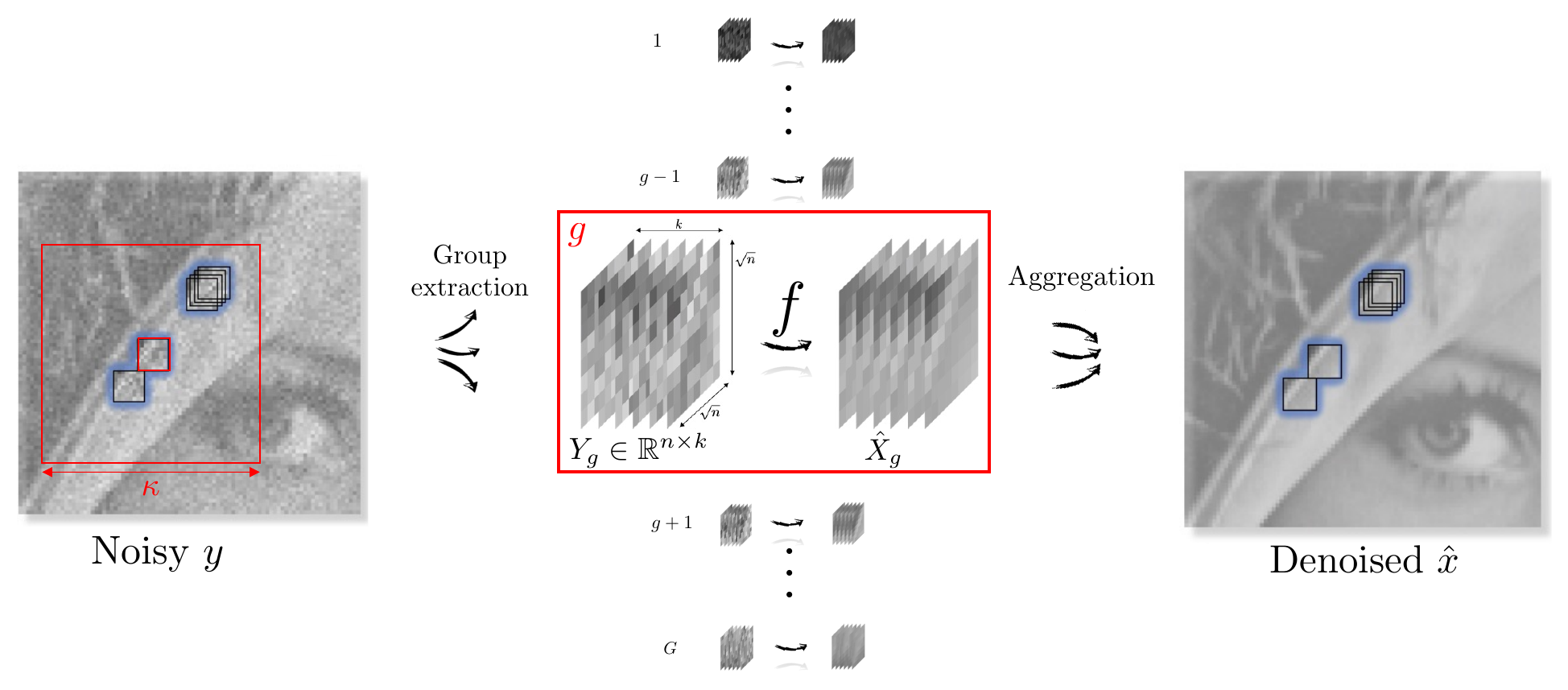} 
\caption[Grouping technique for image denoising]{Illustration of the grouping technique for image denoising.}
\label{bm0_2}
\end{figure}

 BM3D (Block Matching 3D)  \cite{BM3D}  is a powerful and widely acclaimed algorithm that has achieved remarkable success in image denoising. BM3D  considerably improves the performance of pure sparsity methods such as \cite{DCT, wavelet, wavelet_sure} by adding another key element, namely the grouping technique, exploiting the redundancy present in natural images.
 Figure \ref{bm0_2} illustrates this popular technique in image denoising \cite{BM3D, nlbayes, WNNM, LSSC, SAIST, NCSR, PLR}. It basically consists in grouping image patches based on patch resemblance into 3D blocks, also referred to as similarity matrices, in order to perform collaborative filtering.
 During the first stage of BM3D, the denoising of the independent 3D blocks is performed by assuming a local sparse representation in a transform domain. Essentially, BM3D solves the same optimization problem as (\ref{sparsity_optimization}) with the only difference that it involves groups of patches instead of processing each patch separately. Among the possible bases of decomposition for processing the groups, 3D-DCT is frequently used and the same fast procedure as \cite{DCT} that consists in canceling all coefficients below a given threshold is adopted for fast resolution. As for the second stage, Wiener filtering is leveraged for collaborative denoising. 
 Overall, the remarkable denoising performance of BM3D algorithm has made it a widely adopted and benchmark denoising method in applications.

\subsubsection{Sparsity on a learned dictionary}

The use of a fixed basis such as the DCT or the DWT for sparsity-based image denoising has the advantage of being both general and fast. However, it is not easy to know in advance which basis to choose for achieving the best denoising results on a given image, although some attempts have been made in this direction \cite{wavelet_best2, surelet}. A more flexible approach is to directly adapt the decomposition to the input image by unsupervised learning. 

\paragraph{KSVD} KSVD \cite{ksvd} is a popular unsupervised learning algorithm for creating an adaptive dictionary for sparse representations. Formally, let $Y \in \mathbb{R}^{n \times N}$ be the matrix gathering all the $N$ overlapping vectorized patches of size $n$ of a noisy image, $D \in \mathbb{R}^{n \times d}$ an overcomplete dictionary (a set of $d \geq n$ patches, also referred to as atoms, spanning the entire signal space), and $\Theta \in \mathbb{R}^{d \times N}$ the sparse coefficients of the linear combinations. The optimization problem at the heart of KSVD \cite{ksvd} is the following:
\begin{eqnarray}
     \arg \min_{D, \Theta} \| Y - D\Theta \|_F^2 ~\text{s.t.}~  \| \Theta_{\cdot, j} \|_0 \leq t_0, \forall j \in \{1, \ldots, N\},
     \label{ksvd_optimization}
\end{eqnarray}
\noindent where $t_0 \leq n$ is an hyperparameter controlling the sparsity of the linear combinations. Note that this objective is very similar with (\ref{sparsity_optimization}), with the difference that the dictionary $D$ is no longer fixed but fully integrated to the learning process. The resolution of (\ref{ksvd_optimization}) is achieved via an alternating optimization algorithm by iteratively fixing dictionary $D$ and coefficients $\Theta$ as follows:

\noindent \textsc{Sparse coding:} For a dictionary $D$ fixed, solving (\ref{ksvd_optimization}) amounts to solving $N$ independent subproblems for which any pursuit algorithm \cite{matching_pursuit3, matching_pursuit2, matching_pursuit} can be leveraged  for resolution. Indeed, we have:
\begin{equation}
\| Y - D\Theta \|_F^2 = \sum_{j=1}^{N} \| Y_{\cdot, j} - D\Theta_{\cdot, j}\|_2^2\,.
\label{Y-DTheta}
\end{equation}

\noindent \textsc{Dictionary updating:} Assuming that both $D$ and $\Theta$ are fixed, except one column in the dictionary $D_{\cdot, k}$ (atom $k$) and its corresponding 
coefficients $\Theta_{k, \cdot}$, (\ref{Y-DTheta}) can be rewritten as:
\begin{eqnarray}
\| Y \!-\! D\Theta \|_F^2 = \| Y \!-\! \sum_{j=1}^{d} D_{\cdot, j} \Theta_{j, \cdot} \|_F^2  = \left\lVert E_k \!-\! D_{\cdot, k} \Theta_{k, \cdot} \right\rVert_F^2,
\end{eqnarray}
\noindent where $E_k := Y -  \sum_{j \neq k} D_{\cdot, j} \Theta_{j, \cdot}$. In other words, it amounts to finding a matrix of rank $1$ minimizing the $\ell_2$ distance with $E_k$. The solution can be computed using the singular value decomposition (SVD) according to Eckart-Young theorem \cite{eckart}. Formally, let $u$, $v$ and $s$ be the first left-singular vector, right-singular vector and singular value of $E_k$, respectively. Then, its closest matrix of rank $1$, in the $\ell_2$ sense, is simply $s u v^\top$. However, it is very likely that the first right-singular value of $E_k$ is not sparse; therefore, it cannot be used to update coefficients  $\Theta_{j, \cdot}$. The trick of KSVD \cite{ksvd} consists in modifying only the nonzero entries of $\Theta_{j, \cdot}$, thus ensuring that it stays sparse. Mathematically, it comes down to computing the SVD of $E_k$ for which the columns corresponding to a zero coefficient in $\Theta_{j, \cdot}$ have been deleted.

\begin{figure}[t]
    \centering
\includegraphics[width=0.8\columnwidth]{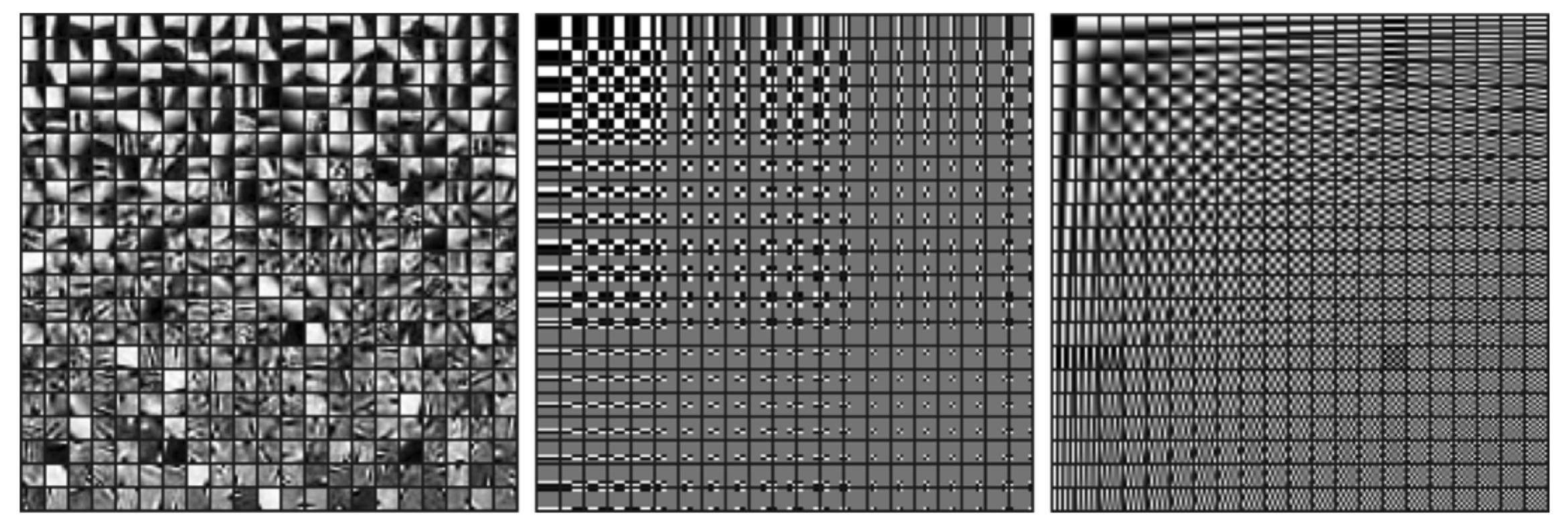}        
    \caption[Data-driven dictionary by KSVD vs overcomplete DCT/Haar dictionaries]{Example of the learned dictionary by KSVD algorithm \cite{ksvd} (left), the overcomplete separable Haar dictionary (middle) and the overcomplete DCT dictionary (right). Source: \cite{ksvd2}.}
    \label{figure_ksvd}
\end{figure}

Following this alternating optimization procedure, KSVD \cite{ksvd} converges after a few iterations. At the end, all denoised patches are repositioned at their initial locations and averaged to produce the final denoised image. Interestingly, the dictionary learned in an unsupervised fashion can be displayed (see Fig. \ref{figure_ksvd}). In spite of its great theoretical interest, KSVD \cite{ksvd} is unfortunately little used in practice, due to its tedious optimization procedure, its difficult-to-set hyperparameters and its limited performance compared to BM3D \cite{BM3D}.

\paragraph{Simultaneous sparse coding (SSC) from a low-rank view point} While KSVD \cite{ksvd} tries to learn a general overcomplete dictionary, for which every patch of the input image can be reconstructed using only a few atoms, some authors argue that the dictionary should be adaptive to groups of similar patches to improve the performance of sparse representation models \cite{LSSC, SAIST, WNNM, PLR}. Indeed, a major drawback of (\ref{ksvd_optimization}) is the assumption about the independence between sparsely-coded patches. In order to better exploit the self-similarity of patches in an image, a refinement consists in constraining the similar patches to share the same atoms in their sparse coding.% (simultaneous sparse coding; see Fig. \ref{figure_ssc}). 
To that end, the optimization problem (\ref{ksvd_optimization}) can be slightly adapted to groups of similar patches, making it even more restricted: 
\begin{equation}
     \arg \min_{D, \Theta} \| Y - D\Theta \|_F^2 \quad \text{s.t.} \quad \| \Theta \|_0 \leq t_0 \,,
     \label{nnksvd_optimization}
\end{equation}
where $Y \in \mathbb{R}^{n \times k}$ is a similarity matrix, $D \in \mathbb{R}^{n \times d}$ a dictionary, $\Theta \in \mathbb{R}^{d \times k}$ the sparse coding, and where the matrix \textit{pseudo} $\ell_0$ norm counts the number of non-zero rows. Note in particular that, subject to dimensional compatibility, any admissible point $\Theta$ for (\ref{nnksvd_optimization}) is also admissible for  (\ref{ksvd_optimization}). Moreover, it is worth noting that the dictionary becomes strictly local under the group sparse representation contrary to (\ref{ksvd_optimization}). As a matter of fact, solving (\ref{nnksvd_optimization}) amounts to solving a low-rank approximation of $Y$ if we denote  $X = D\Theta$:
\begin{equation}
    \arg \min_X  \| Y - X \|_F^2  \quad \text{s.t.} \quad \operatorname{rank}(X) \leq t_0\,,
    \label{nnksvd_optimization_rank}
\end{equation}
for which the solution is expressed with the help of the singular value decompostion (SVD) of $Y$ according to Eckart-Young theorem \cite{eckart}. In particular, considering the Lagrangian unconstrained formulation  of (\ref{nnksvd_optimization_rank}) with hyperparameter $\gamma \geq 0$, we have (see proof in \cite{rank_unconstrained, PLR}):
\begin{equation}
    U \varphi_{\text{hard}, \sqrt{\gamma}}(S) V^\top = \arg \min_X  \| Y - X \|_F^2 + \gamma \operatorname{rank}(X)\,,
    \label{nnksvd_optimization_rank2}
\end{equation}
where $Y = U S V^\top$ is the SVD of $Y$ and $\varphi_{\text{hard}, \gamma}$ denotes the hard shrinkage  operator that applies element-wise $\varphi_{\text{hard}, \gamma}(x) =
 x \mathbf{1}_{\mathbb{R} \backslash [-\gamma, \gamma]}(x)$. Equation (\ref{nnksvd_optimization_rank2}) is at the core of PLR algorithm \cite{PLR} where the value of $\gamma = 2.25\, k\, \sigma^2$ is recommended experimentally for denoising  $Y$ when it is corrupted by additive white Gaussian noise (AWGN) of variance $\sigma^2$.

A relaxation of (\ref{nnksvd_optimization}) is proposed by LSSC \cite{LSSC} through a so-called grouped-sparsity regularizer to encourage the alignment of sparse coefficients along the row direction, without imposing it as a hard constraint. Interestingly, this relaxation has also a low-rank interpretation  from a variance estimation perspective \cite{SAIST}. Specifically, it amounts to solving a nuclear norm minimization problem (NNM) which has a closed-form solution \cite{NNM}:   
\begin{equation}
   U \varphi_{\text{soft}, \gamma}(S) V^\top = \arg \min_X  \frac{1}{2} \| Y - X \|_F^2 + \gamma \| X \|_\ast\,,
   \label{nnm_problem}
\end{equation}
\noindent  where $\| X \|_\ast$ denotes nuclear norm (sum of the singular values) and $\varphi_{\text{soft}, \gamma}$ denotes the soft shrinkage operator that applies element-wise $\varphi_{\text{soft}, \gamma}(x) = \operatorname{sign}(x) \cdot \max(|x| - \gamma, 0)$. More recently, WNNM \cite{WNNM} refines the NNM problem (\ref{nnm_problem}) by assigning different weights to the singular values. Combined with iterative regularization technique \cite{iterative_regularization} and  multiple passes of denoising, WNNM  achieves state-of-the-art performances.

%------------------------------------------------------------------------------------------------------------------------------------
% SECTION 7
%------------------------------------------------------------------------------------------------------------------------------------

\subsection{Bayesian methods coupled with a Gaussian model}

Bayesian methods coupled with a Gaussian model form a powerful framework for probabilistic modeling in image denoising \cite{hyperprior, EPLL, nlbayes, EPLL_unsupervised, PLE, noise_clinic}. The Gaussian model (or a mixture of Gaussians)  is widely used due to its simplicity and flexibility in capturing a wide range of continuous data, signal being no exception. In this section, we review two algorithms \cite{EPLL, nlbayes} that are major representatives of Bayesian modeling under Gaussian prior. 

\paragraph{EPLL} Based on the observation that learning good image priors over whole images is challenging, EPLL \cite{EPLL} proposes to transpose the modeling of an image prior back to the prior modeling of small image patches, which is assumed to be an easier task. Specifically, in the case of additive white Gaussian noise (AWGN) of variance $\sigma^2$, the maximum a posteriori (MAP) of the clean patch $x_i$ given its noisy patch $y_i \sim \mathcal{N}(x, \sigma^2 I_n)$ and an arbitrary image patch prior $p$ leads to the minimization of an  energy (via Bayes' rule):
\begin{equation}
    E_i(x_i, y_i) := \frac{1}{2\sigma^2} \| x_i - y_i \|_2^2 - \log p(x_i)\,.
\end{equation}

\noindent In order to extend this energy to the whole image, EPLL \cite{EPLL_IPOL} defines the global energy of a full image $x$, given $y$, by the average energy of energies all its $N$ overlapping patches: 
\begin{equation}
\begin{aligned}
    E(x, y) := \frac{1}{N} \sum_{i=1}^{N} E_i(x_i, y_i) 
     \approx \frac{n}{2\sigma^2 N} \|x -y \|_2^2 - \text{EPLL}_p(x) \,,
\end{aligned}
\end{equation}
\noindent where $\text{EPLL}_p(x) :=  \sum_{i}  \log p(x_i) / N$ is the expected patch log likelihood (EPLL). Note that the approximation is legitimate because a noisy pixel belongs to exactly $n$ overlapping patches, with the exception of pixels located close to the borders, which are neglected for the sake of simplicity. Searching for the image $x$ with smallest global energy, the resulting optimization problem finally reads \cite{EPLL}:
\begin{eqnarray}
\arg \min_x \frac{n}{2 N \sigma^2} \|x -y \|_2^2 - \text{EPLL}_p(x)\,.
\label{epll_equation}
\end{eqnarray} 
Thus, the expected patch log likelihood (EPLL) acts as a regularization term in  (\ref{epll_equation}). Direct optimization of the cost function (\ref{epll_equation}) may be very hard, depending on the prior used. That is why, half quadratic splitting \cite{GemanYang02} is leveraged for efficient resolution by introducing auxiliary variables. Note that this iterative optimization method shares close links with the alternating direction method of multipliers (ADMM) \cite{BoydADMM11}.

Although this framework allows the use of patch priors $p$ of any sort, the authors \cite{EPLL} propose to leverage a surprisingly simple Gaussian mixture model: $p(x) = \sum_{k=1}^K \pi_k \mathcal{N}(x; \mu_k, \Sigma_k),$
\noindent where $\pi_k$ are the mixing weights for each of  the mixture component and the $\mu_k$ and $\Sigma_k$ are the corresponding mean and covariance matrix. In the original paper \cite{EPLL}, the parameters $\pi_k, \mu_k$ and $\Sigma_k$ for $K=200$ components are estimated in a supervised fashion by maximum likelihood estimation (MLE) over a set of two millions clean patches collected from BSD dataset \cite{berkeley} using the expectation–maximization (EM) algorithm for optimization. However, a recent work \cite{EPLL_unsupervised} shows that Gaussian mixture parameters can also be estimated unsupervisedly directly from the noisy input image itself, resulting in even better image denoising performance than its supervised counterpart.

Despite their significant theoretical relevance, EPLL \cite{EPLL} and its variants \cite{EPLL_unsupervised, deledalle2018image} have not gained much popularity in practical applications primarily due to their cumbersome optimization procedures when compared to the well-established BM3D \cite{BM3D} algorithm. Finally, note that the EPLL approach \cite{EPLL} shares some similarities with the Field-of-Experts framework \cite{FoE} designed  previously where the parameterized density function, namely the Gaussian mixture model, is replaced by a Product-of-Experts \cite{PoE} that exploits non-linear functions of many linear filter responses and where optimization is essentially performed through gradient ascent.

\iffalse
Based on the observation that learning good image priors over whole images is challenging, EPLL \cite{EPLL} proposes to transpose the modeling of an image prior back to the  modeling of small image patches, which is assumed to be an easier task. Specifically, for a clean image $x \in \mathbb{R}^n$ which decomposes into $N$ overlapping small image patches $\{x_i\}_{i=1, \ldots, N}$, the proposed arbitrary image prior $p$ is as follows:
\begin{equation}
    p(x) \propto \prod_{i=1}^{N} p(x_i)\,.
\end{equation} 
\noindent In other words, an image is considered likely if and only if all overlapping image patches are likely under the patch prior. It should be noted that a simpler model would have consisted in taking into account only non-overlapping patches, which would be tantamount to considering all patches as independent. However, it is well-known that such a model induces notorious artifacts at patch borders.

Assuming that an image is corrupted by additive white Gaussian noise (AWGN) of variance $\sigma^2$, the noise model reads $p(y|x) = \mathcal{N}(x, \sigma^2 I_n)$. Applying Bayes' rule yields the following posterior probability:
\begin{equation}
\underbrace{p(x|y)}_{\text{posterior}} \propto \underbrace{p(y|x)}_{\text{likelihood}}  \times \underbrace{p(x)}_{\text{prior}}\,.
\end{equation} 
argmax = argmin...
Note that equation introduced this way in the original paper
\fi

\paragraph{NL-Bayes:} The N(on)-L(ocal) Bayes  \cite{nlbayes} algorithm combines the concepts of Bayesian modeling and self-similarity \cite{irani} which appears to be key element for achieving state-of-the-art results. Formally, let $y \sim \mathcal{N}(x, \sigma^2 I_n)$ be a noisy image patch corrupted by Gaussian noise of variance $\sigma^2$. Arbitrarily setting a multivariate Gaussian prior on the clean patch $x$, \textit{i.e.} $p(x) = \mathcal{N}(x; \mu, \Sigma)$, where the mean $\mu$  and covariance matrix $\Sigma$ are to be estimated,  the maximum a posteriori (MAP), computed using Bayes’ rule, is the solution of the following optimization problem:
\begin{eqnarray}
\begin{aligned}
    \hat{x}_{MAP} &= \arg \max_{x} \frac{1}{2\sigma^2} \|x - y\|_2^2 + \frac{1}{2} (x-\mu)^\top \Sigma^{-1} (x-\mu)\\
    & = \mu +  \Sigma \left(\Sigma + \sigma^2 I_n \right)^{-1}  (y - \mu) \,.
\end{aligned}
\label{nlbayes_key_equation}
\end{eqnarray}

\noindent NL-Bayes \cite{nlbayes} proposes to construct the prior $p(x)$ from the group of image patches similar to $x$.  Specifically, let $X \in \mathbb{R}^{n \times k}$ be the similarity matrix of $x$. Then, $\mu$ is estimated by the average patch and $\Sigma$ is estimated by the empirical covariance matrix of the group, that is:
\begin{equation}
    \mu_X := \frac{1}{k} X \mathbf{1}_k ~ \text{and} ~ \Sigma_X := \frac{1}{k} (X - \mu_X \mathbf{1}_k^\top) (X - \mu_X \mathbf{1}_k^\top)^\top.
    \label{nl_bayes_mu_sigma}
\end{equation}
\noindent But of course, the true similarity matrix $X$ is unknown and can only be deduced from its noisy version $Y$. To that end, unbiased estimates are leveraged as a first approximation, namely $ \mu_Y = \frac{1}{k} Y \mathbf{1}_k $ and $\Sigma_Y = \frac{1}{k} (Y - \mu_Y \mathbf{1}_k^\top) (Y - \mu_Y \mathbf{1}_k^\top)^\top - \sigma^2 I_n$. Using equation (\ref{nlbayes_key_equation}) and the two estimates $\mu_Y$ and $\Sigma_Y$, each noisy patch of a given similarity matrix can then be denoised. Viewing this denoising step as a function $f$ that processes similarity matrices, NL-Bayes falls into the category of non-local denoisers (see Fig. \ref{bm0_2}). 

After reprojection and aggregation by average of all patch estimates, a first denoised image is built which is exploited to refine the priors $p(x)$ for each group of similar patches. Actually, using the equation (\ref{nl_bayes_mu_sigma}) with approximate similarity matrices gives better results in practice than with the unbiased estimates of the first step. 
Repeating this second stage again and again, taking advantage of the availability of a supposedly better image estimate than in the
previous step, does not bring experimentally any improvements unfortunately. That is why, the algorithm stops after two steps.

NL-Bayes \cite{nlbayes} compares favorably with BM3D \cite{BM3D} and is as competitive as in terms of speed which makes it an interesting alternative for practical image denoising \cite{noise_clinic}.

%------------------------------------------------------------------------------------------------------------------------------------
% SECTION 8
%------------------------------------------------------------------------------------------------------------------------------------

\subsection{Deep learning-based methods}

In recent years, attempts have been made to reconcile unsupervised learning and deep neural networks in image denoising \cite{S2S, DIP, N2S}. Major representatives are Deep Image Prior (DIP) \cite{DIP} and Self2self \cite{S2S}.

\paragraph{Deep Image Prior}
Deep Image Prior \cite{DIP} adopts a non-intuitive strategy which consists in training a convolutional neural network with U-net architecture $f_\theta$ to predict the input noisy image $y \in \mathbb{R}^n$ from a single realization of pure uniform noise $u \sim \mathcal{U}([0, 1]^{n \times C})$, where $C$ denotes the number of feature maps (\textit{e.g.} $C = 32$):
\begin{equation}
     \arg \min_\theta  \| f_\theta(u) - y \|_2^2 \,.
     \label{dip_equation}
\end{equation}
By early stopping the optimization process based on gradient descent in order to avoid perfect reconstruction of the noisy image $y$, it is observed that $f_{\theta}(u)$ may be surprisingly very close to the true image $x$ in practice. According to the authors \cite{DIP}, this intriguing phenomenon is an evidence that realistic images are naturally promoted by certain types of neural networks. Indeed, equation (\ref{dip_equation}) is only composed of the data-fidelity term without any regularizer, which suggests that network architectures actually encode an implicit image prior. 

Some refinements of (\ref{dip_equation}) were proposed afterwards to enhance the performance of DIP \cite{DIP} by adding nevertheless an explicit prior. For example, combining DIP \cite{DIP} with the traditional TV regularization \cite{TV} was investigated in \cite{DIP_TV} with relative quality improvement, at the price of an extra hyperparameter  balancing the data-fidelity term and the regularization term. Another interesting alternative \cite{DIP_RED} consists in explicitly regularizing DIP \cite{DIP} using an existing unsupervised denoising algorithm such as BM3D \cite{BM3D}. The concept of regularization by denoising (RED) \cite{RED} is indeed an alternative to Plug-and-Play Prior \cite{PnP} which enables to harness the implicit prior learned by a denoiser to any data-fidelity term, while avoiding the need to differentiate the chosen denoiser. Other variants of DIP \cite{DIP} for improved performance include \cite{dip_rita, rethinking, dip_bayes, subdip}.

\paragraph{Self2Self}
More recently, Self2Self \cite{S2S} considers the pretext task of inpainting to tackle image denoising, namely:
\begin{equation}
    \theta^\ast = \arg \min_\theta \mathbb{E}_b \| (1-b) \odot (f_\theta(b \odot y) - y) \|_2^2 \,,
    \label{slef2self}
\end{equation}
\noindent where $\odot$ denotes the Hadamard product and $b \in \{0, 1\}^n$ is a random vector whose components follow independent Bernoulli distributions with probability $p \in (0, 1)$. This time, droupout \cite{dropout} and sampling are exploited as regularization techniques for avoiding the convergence to the constant function $f_\theta(.) = y$. During the inference step, about a hundred of artificially simulated samples $f_{\theta^\ast}(b \odot y)$  are averaged for final estimation. Note that
Self2Self \cite{S2S} shares some similarities with Noise2Self \cite{N2S} as $f_\theta$ is learned following the blind-spot strategy. To our knowledge, Self2Self \cite{S2S} is the current state-of-the-art unsupervised deep learning-based denoiser.  

Although promising, the aforementioned deep unsupervised learning methods are still limited in terms of performance and especially in terms of computational cost compared to the patch-based and non-local methods \cite{BM3D, nlbayes, nlridge, WNNM, SAIST, NCSR, PEWA, OWF}. Indeed, deep learning-based methods use the time-consuming gradient descent algorithm for optimization, whereas traditional ones have in general closed-form solutions, which speeds up learning.

%------------------------------------------------------------------------------------------------------------------------------------
% SECTION 9
%------------------------------------------------------------------------------------------------------------------------------------

\section{Normalization-equivariance property of denoising methods}
In many information processing systems, it may be desirable to ensure that any change of the input, whether by shifting or scaling, results in a corresponding change in the system response. While deep neural networks are gradually replacing all traditional automatic processing methods, they surprisingly do not guarantee such normalization-equivariance (scale and shift) property, which can be detrimental in many applications. 

Sometimes wrongly confused with the invariance property which designates the characteristic of a function $f$ not to be affected by a specific transformation $\mathcal{T}$ applied beforehand, the equivariance property, on the other hand, means that $f$ reacts in accordance with $\mathcal{T}$. Formally, invariance is $f \circ \mathcal{T} = f$ whereas equivariance reads $f \circ \mathcal{T} = \mathcal{T} \circ f$, where $\circ$ denotes the function composition operator. 
Both invariance and equivariance play a crucial role in many areas of study, including computer vision and signal processing and have recently been studied in various settings for deep-learning-based models \cite{equivariant1, equivariant_graph2, equivariant_graph3, equivariant_graph4, equivariant_image, equivariant_image2, equivariant_image3, equivariant_image4, equivariant_pointcloud1, equivariant_pointcloud2, equivariant_pointcloud3, equivariant_translation1, equivariant_translation2}. 

In this section, we focus on the equivariance of superivised and unsupervised image denoising methods  $f_\theta$ to a specific transformation $\mathcal{T}$, namely normalization. 

%------------------------------------------------------------------------------------------------------------------------------------
% SECTION 10
%------------------------------------------------------------------------------------------------------------------------------------

\subsection{Normalization-equivariance of conventional denoisers} 

We start with formal definitions of the  different types of equivariances studied in this chapter. \textcolor{black}{Please note that our definition of ``scale'' and ``shift'' may differ from the definition given by some authors in the image processing literature.}

\begin{definition} 
A function $f : \mathbb{R}^n \mapsto \mathbb{R}^m$ is said to be:

\begin{itemize}
\item \textit{$\mathcal{T}$-equivariant} if $f \circ \mathcal{T} = \mathcal{T} \circ f,$
\item \textit{scale-equivariant} if $\forall x \in \mathbb{R}^n, \forall a \in \mathbb{R}^+_\ast, \: f(ax) = a f(x),
$
\item \textit{shift-equivariant} if  $\forall x \in \mathbb{R}^n, \forall b \in \mathbb{R}, \:  f(x + b) = f(x) + b,$
\item \textit{normalization-equivariant} if it is both \textit{scale-equivariant} and \textit{shift-equivariant}: 
$$
\forall x \in \mathbb{R}^n, \forall a \in \mathbb{R}^+_\ast, \forall b \in \mathbb{R}, \: f(ax + b) = af(x) + b, 
$$
\end{itemize}
\noindent where addition with the scalar shift $b$ is applied element-wise.
\label{normalization-equivariance}
\end{definition}
Note that the \textit{scale-equivariance} property is more often referred to as positive homogeneity in pure mathematics. S. Mohan et al. \cite{homogeneous_functions}   revealed that scale-equivariant neural networks could simply be built by removing the additive constant ("bias") terms in CNNs with ReLU activation functions without affecting performance. Moreover, they showed that a much better generalization at noise levels outside the training range was ensured by these networks as a spectacular outcome.

A (``blind'') denoiser is basically a function $f : \mathbb{R}^n \mapsto \mathbb{R}^n$ which, given a noisy image $y \in \mathbb{R}^n$, tries to map the corresponding noise-free image $x \in \mathbb{R}^n$. Since scaling up an image by a positive factor $a$ or adding it up a constant shift  $b$ does not change its contents,  it is natural to expect scale and shift equivariance, \textit{i.e.} normalization equivariance, from the denoising procedure emulated by $f$.

The most basic methods for image denoising are the smoothing filters (see Section III.A).
It turns out that these ``blind'' denoisers process images by convex combinations of pixels and therefore all implement a \textit{normalization-equivariant} function. More generally, one can prove that a linear filter is \textit{normalization-equivariant} if and only if its coefficients add up to $1$.  In others words, \textit{normalization-equivariant} linear filters process images by affine combinations of pixels.
As such, $f_{\operatorname{NLM}} = y^\top \theta$ (with the weights $\theta$ given in (\ref{nlm_weights})) is a \textit{normalization-equivariant} function. More recently, NL-Ridge \cite{nlridge} and LIChI \cite{lichi} propose to process images by linear combinations of similar patches and achieves state-of-the-art performance in unsupervised denoising. When restricting the coefficients of the combinations to sum to $1$, that is imposing affine combination constraints, the resulting algorithms encode  \textit{normalization-equivariant} functions as well.

\addtolength{\tabcolsep}{-1.5pt} 
\begin{table}[t]
  \caption[Equivariance properties of several image denoisers]{Equivariance properties of several image denoisers (left: traditional, right: deep learning-based)}
  \label{properties_summary}

  \centering
  \begin{tabular}{cccccccc}
  & \multicolumn{7}{c}{\it  Traditional}  \\
         & TV      & NLM  & NL-Ridge & LIChI & DCT & BM3D &  WNNM  \\
     \hline
    Scale-eq & \textcolor{greenG}{\ding{51}}
     &  \textcolor{greenG}{\ding{51}}
     &  \textcolor{greenG}{\ding{51}}  
     &  \textcolor{greenG}{\ding{51}} &  \textcolor{greenG}{\ding{51}}  & \textcolor{greenG}{\ding{51}} & \textcolor{greenG}{\ding{51}}
      \\
    Shift-eq &  \textcolor{greenG}{\ding{51}}
     &   \textcolor{greenG}{\ding{51}}
     & \textcolor{greenG}{\ding{51}} & \textcolor{greenG}{\ding{51}}
     &  \textcolor{red}{\ding{55}}  & \textcolor{red}{\ding{55}} & \textcolor{red}{\ding{55}} 
    \\
    \hline
  \end{tabular}

\bigskip 
   \begin{tabular}{ccccccc}
  & \multicolumn{6}{c}{\it Deep-learning} \\
        &  DnCNN  & NLRN & SCUNet & Restormer & DCT2net & DRUNet \\
     \hline
    Scale-eq & \textcolor{red}{\ding{55}}
     &  \textcolor{red}{\ding{55}}  & \textcolor{red}{\ding{55}} &  \textcolor{red}{\ding{55}} & \textcolor{greenG}{\ding{51}} & \textcolor{greenG}{\ding{51}} \\
    Shift-eq & \textcolor{red}{\ding{55}}
       & \textcolor{red}{\ding{55}} & \textcolor{red}{\ding{55}} & \textcolor{red}{\ding{55}} & \textcolor{red}{\ding{55}} & \textcolor{red}{\ding{55}} \\
    \hline\\
  \end{tabular}
\end{table}

\begin{figure}[t]
    \centering
\includegraphics[width=0.9\columnwidth]{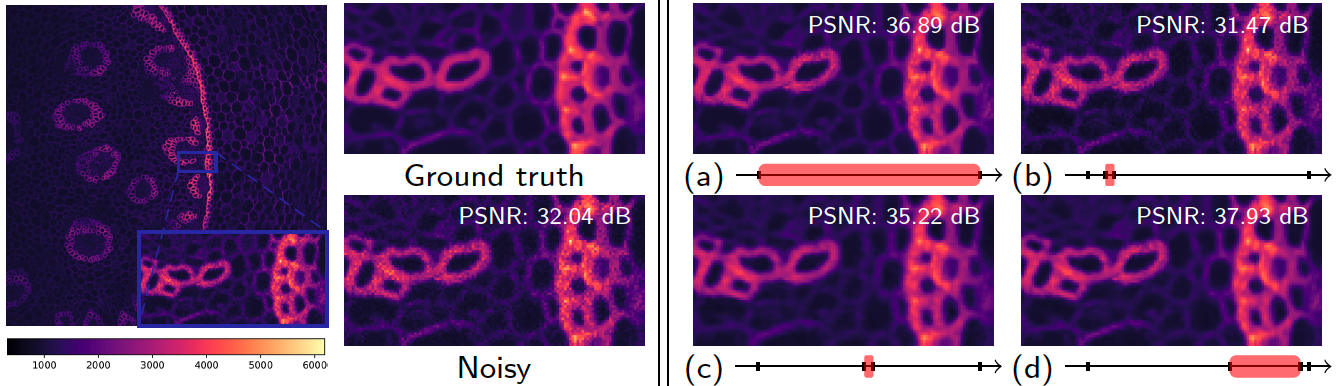}\\ 
\caption{Impact of normalization for deep-learning-based denoising (DnCNN). All four intervals are subsets of the interval $[0,1]$. Source: \cite{SHneurips23}.}      
    \label{FigEquivData}
\end{figure}

\subsection{The case of neural networks}

Deep learning hides a subtlety about normalization equivariance that deserves to be highlighted. Usually, the weights of neural networks are learned on a training set containing data all normalized to the same arbitrary interval $[s_0, t_0]$. 
At inference, unseen data are processed within the interval $[s_0, t_0]$ via a $s\hbox{-}t$ linear normalization with $s_0 \leq s < t \leq t_0$ denoted $\mathcal{T}_{s, t}$  and defined by:
\begin{equation}
 \mathcal{T}_{s , t}: y \mapsto (t-s)\frac{y - \min(y)}{\max(y) - \min(y)} + s\,.
\end{equation}
Note that this transform is actually the unique linear one with positive slope that exactly bounds the output to $[s, t]$. The data is then passed to the trained network and its response is finally returned to the original range via the inverse  operator $\mathcal{T}^{-1}_{s, t}$. This proven pipeline is actually relevant in light of the following proposition.

\begin{proposition}
     $\forall \: s < t \in \mathbb{R}, \forall \: f :  \mathbb{R}^n \mapsto \mathbb{R}^m,
 \mathcal{T}^{-1}_{s, t} \circ f \circ \mathcal{T}_{s, t}$ is a \textit{normalization-equivariant} function.
 \label{case_of_nn}
\end{proposition}

Thus, if $f$ is not normalization-equivariant, $\mathcal{T}_{s, t}^{-1} \circ f \circ \mathcal{T}_{s, t}$ is guaranteed to be.
While normalization-equivariance appears to be solved, a question is still remaining: how to choose the hyperparameters $s$ and $t$ for a given function $f$ ? 
Obviously, a natural choice for neural networks is to take the same parameters $s$ and $t$ as in the learning phase whatever the input image is, \textit{i.e.} $s=s_0$ and $t=t_0$, but are they really optimal ? The answer to this question is generally negative. 
It is particularly true in image denoising when the noise level at inference differs from the noise level on which the neural network was trained on. 
Indeed, as showed by \cite{normalization_morel}, careful choices on parameters $s > s_0$ and $t < t_0$ can strongly mitigate the effect of the distribution gap (see Figure \ref{FigEquivData} and Figure 5 in \cite{SHneurips23}).
This suggests that there are always inherent performance leaks for deep neural networks due to the two degrees of freedom induced by the normalization (\textit{i.e.}, choice of $s$ and choice of $t$). In addition, this poor conditioning can be a source of misinterpretation in critical applications.

\subsection{Categorizing image denoisers}
Table \ref{properties_summary} summarizes the equivariance properties of several popular denoisers, either conventional \cite{TV, nlmeans, nlridge, lichi, DCT, BM3D, WNNM} or deep-learning-based \cite{dncnn, drunet, nlrn, swinir}. Interestingly, if \textit{scale-equivariance} is generally guaranteed for traditional denoisers, not all of them are equivariant to shifts. In particular, the widely used algorithms DCT \cite{DCT} and BM3D \cite{BM3D} are sensitive to offsets, mainly because the hard thresholding function at their core is not \textit{shift-equivariant}. 

\begin{figure}[t]
    \centering
\includegraphics[width=0.9\columnwidth]{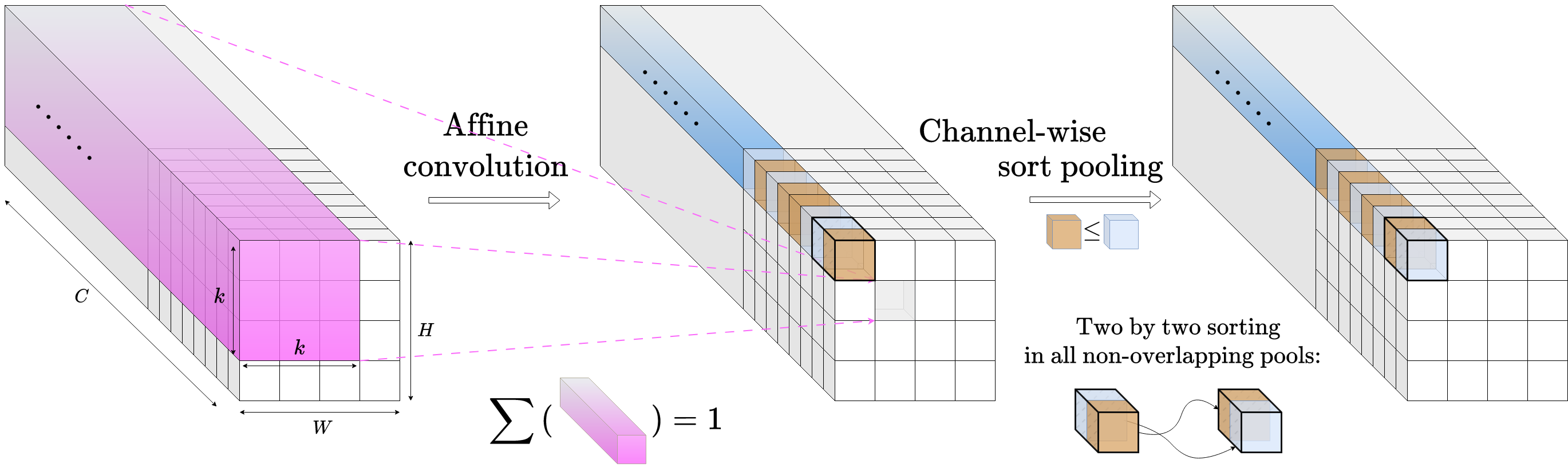}
    \caption[Proposed methodology for designing normalization-equivariant neural networks]
	{Illustration of the alternative for replacing the traditional scheme ``convolution + element-wise activation function'' in CNNs:
    affine convolutions supersede ordinary ones by restricting the coefficients of each kernel to sum to one and the proposed sort pooling patterns introduce nonlinearities by sorting two by two the pre-activated neurons along the channels. Source: \cite{SHneurips23}. }
    \label{sortpooling_intro}
\end{figure}

Regarding the deep-learning-based networks, only DRUNet \cite{drunet} is insensitive to scale because it is a bias-free convolutional neural network with only ReLU activation functions \cite{homogeneous_functions}. \textcolor{black}{In particular, all transformer models \cite{restormer, n3net, nlrn, swinir, transformer_eccv, scunet}, even bias-free, are not \textit{scale-equivariant}  due to their inherent attention-based modules.}  To solve this issue,  an approach was proposed in \cite{SHneurips23} to adapt the architecture of existing neural networks so that normalization equivariance
holds by design.
The authors argue that the classical pattern "conv+ReLU" can be favorably replaced affine convolutions that ensure that all coefficients of the convolutional kernels sum to one in one hand, and the other hand, channel-wise sort pooling nonlinearities as a substitute for all activation functions that apply element-wise, including ReLU or sigmoid functions  (see Fig. \ref{sortpooling_intro}). 
These two architectural modifications do preserve normalization-equivariance without loss of performance,  
and provide much better generalization across noise levels in practice because they can naturally adapt to the best-performing interval \cite{SHneurips23} (see Figure \ref{FigEquivData} and Figure 5 in \cite{SHneurips23}).

\section{Conclusion}

In this paper, we have surveyed supervised and unsupervised image denoising methods, categorized the methods based on their equivariance properties, and summarized quantitatively their performances assessed on popular benchmarks, such as Set12 and BSD68.
It is clearly established that the traditional unsupervised methods \cite{BM3D, nlbayes, WNNM, lichi} outperform the deep-learning counterparts, so far. 
They are particularly recommended when it is not possible to collect enough high-quality dataset for training denoising models or too much time consuming.
In return, it is also clear that supervised methods based on CNNs boosted  with attention modules \cite{scunet} surpass top-rank traditional methods \cite{WNNM, lichi} (see Fig. \ref{PSNRcpu}), capable of recovering details barely perceptible by an human.      
The question is which architecture will prevail over the next ten years. Transformer-based methods show great promise for image denoising and are likely to play an important role in the future. 
Meanwhile, let us not bury too quickly the other architectures as comebacks are still possible \cite{convnext, convmixer}. 
The potential of all these frameworks has not yet been fully explored and open challenges and promising research directions for image denoising in the coming years.
Neverthelss, in view of the recent spectacular results, it is legitimate to wonder whether we are approaching the theoretical limit of denoising performance, which could reopen the debate on whether image denoising is close to death \cite{dead, dead2, dead3}

\bibliographystyle{abbrv} 
\bibliography{bib}

\begin{thebibliography}{100}

\bibitem{fully_synthetic_training}
R.~Achddou, Y.~Gousseau, and S.~Ladjal.
\newblock Fully synthetic training for image restoration tasks.
\newblock {\em Computer Vision and Image Understanding}, 233:103723, 2023.

\bibitem{hyperprior}
C.~Aguerrebere, A.~Almansa, Y.~Gousseau, J.~Delon, and P.~Mus{\'e}.
\newblock A hyperprior {B}ayesian approach for solving image inverse problems.
\newblock {\em HAL preprint}, 2014.

\bibitem{dset_div2k}
E.~Agustsson and R.~Timofte.
\newblock {NTIRE 2017 Challenge} on single image super-resolution: Dataset and study.
\newblock In {\em Conference on Computer Vision and Pattern Recognition Workshops (CVPRW)}, pages 1122--1131, 2017.

\bibitem{ksvd2}
M.~Aharon, M.~Elad, and A.~Bruckstein.
\newblock {K-SVD}: An algorithm for designing overcomplete dictionaries for sparse representation.
\newblock {\em IEEE Transactions on Signal Processing}, 54(11):4311--4322, 2006.

\bibitem{ridnet}
S.~Anwar and N.~Barnes.
\newblock Real image denoising with feature attention.
\newblock In {\em International Conference on Computer Vision (ICCV)}, October 2019.

\bibitem{N2S}
J.~Batson and L.~Royer.
\newblock {N}oise2{S}elf: Blind denoising by self-supervision.
\newblock In {\em International Conference on Machine Learning (ICML)}, volume~97, pages 524--533, 2019.

\bibitem{equivariant_graph4}
S.~Batzner, A.~Musaelian, L.~Sun, M.~Geiger, J.~P. Mailoa, M.~Kornbluth, N.~Molinari, T.~E. Smidt, and B.~Kozinsky.
\newblock E(3)-equivariant graph neural networks for data-efficient and accurate interatomic potentials.
\newblock {\em Nature Communications}, 13(1):2453, 2022.

\bibitem{surelet}
T.~Blu and F.~Luisier.
\newblock {The SURE-LET approach to image denoising}.
\newblock {\em IEEE Transactions on Image Processing}, 16(11):2778--2786, 2007.

\bibitem{BoulangerTMI2010}
J.~Boulanger, C.~Kervrann, P.~Bouthemy, P.~Elbau, J.-B. Sibarita, and J.~Salamero.
\newblock Patch-based nonlocal functional for denoising fluorescence microscopy image sequences.
\newblock {\em IEEE Transactions on Medical Imaging}, 29(2):442--454, 2009.

\bibitem{BoydADMM11}
S.~Boyd, N.~Parikh, E.~Chu, B.~Peleato, and J.~Eckstein.
\newblock {\em Distributed Optimization and Statistical Learning via the Alternating Direction Method of Multipliers}.
\newblock Now Foundations and Trends, 2011.

\bibitem{unprocessing}
T.~Brooks, B.~Mildenhall, T.~Xue, J.~Chen, D.~Sharlet, and J.~T. Barron.
\newblock Unprocessing images for learned raw denoising.
\newblock In {\em Conference on Computer Vision and Pattern Recognition (CVPR)}, pages 11028--11037, 2019.

\bibitem{nlmeans2}
A.~Buades, B.~Coll, and J.-M. Morel.
\newblock A non-local algorithm for image denoising.
\newblock In {\em Conference on Computer Vision and Pattern Recognition (CVPR)}, volume~2, pages 60--65, 2005.

\bibitem{nlmeans}
A.~Buades, B.~Coll, and J.-M. Morel.
\newblock A review of image denoising algorithms, with a new one.
\newblock {\em Multiscale Modeling \& Simulation}, 4(2):490--530, 2005.

\bibitem{mlp}
H.~C. Burger, C.~J. Schuler, and S.~Harmeling.
\newblock Image denoising: Can plain neural networks compete with {BM3D}?
\newblock In {\em Conference on Computer Vision and Pattern Recognition (CVPR)}, pages 2392--2399, 2012.

\bibitem{equivariant_pointcloud3}
G.~Bökman, F.~Kahla, and A.~Flinth.
\newblock {ZZ-Net}: A universal rotation equivariant architecture for {2D} point clouds.
\newblock In {\em Conference on Computer Vision and Pattern Recognition (CVPR)}, pages 10966--10975, 2022.

\bibitem{NNM}
J.-F. Cai, E.~J. Cand\`{e}s, and Z.~Shen.
\newblock A singular value thresholding algorithm for matrix completion.
\newblock {\em SIAM Journal on Optimization}, 20(4):1956--1982, 2010.

\bibitem{PNGAN}
Y.~Cai, X.~Hu, H.~Wang, Y.~Zhang, H.~Pfister, and D.~Wei.
\newblock Learning to generate realistic noisy images via pixel-level noise-aware adversarial training.
\newblock In {\em Advances in Neural Information Processing Systems (NeurIPS)}, volume~34, pages 3259--3270, 2021.

\bibitem{noise2noise_improved}
A.~F. Calvarons.
\newblock Improved {Noise2Noise} denoising with limited data.
\newblock In {\em Conference on Computer Vision and Pattern Recognition Workshops (CVPRW)}, pages 796--805, 2021.

\bibitem{canny}
J.~Canny.
\newblock A computational approach to edge detection.
\newblock {\em IEEE Transactions on Pattern Analysis and Machine Intelligence}, 8(6):679--698, 1986.

\bibitem{wavelet_christine}
V.~Chappelier and C.~Guillemot.
\newblock Oriented wavelet transform for image compression and denoising.
\newblock {\em IEEE Transactions on Image Processing}, 15(10):2892--2903, 2006.

\bibitem{dead}
P.~Chatterjee and P.~Milanfar.
\newblock Is denoising dead?
\newblock {\em IEEE Transactions on Image Processing}, 19(4):895--911, 2010.

\bibitem{transformer_eccv}
L.~Chen, X.~Chu, X.~Zhang, and J.~Sun.
\newblock Simple baselines for image restoration.
\newblock In {\em European Conference on Computer Vision (ECCV)}, pages 17--33, 2022.

\bibitem{matching_pursuit3}
S.~S. Chen, D.~L. Donoho, and M.~A. Saunders.
\newblock Atomic decomposition by basis pursuit.
\newblock {\em SIAM Journal on Scientific Computing}, 20(1):33--61, 1998.

\bibitem{fast_DCT}
W.-H. Chen, C.~Smith, and S.~Fralick.
\newblock A fast computational algorithm for the {D}iscrete {C}osine {T}ransform.
\newblock {\em IEEE Transactions on Communications}, 25(9):1004--1009, 1977.

\bibitem{tnrd}
Y.~Chen and T.~Pock.
\newblock Trainable nonlinear reaction diffusion: A flexible framework for fast and effective image restoration.
\newblock {\em IEEE Transactions on Pattern Analysis and Machine Intelligence}, 39(6):1256--1272, 2017.

\bibitem{dip_bayes}
Z.~Cheng, M.~Gadelha, S.~Maji, and D.~Sheldon.
\newblock A {B}ayesian perspective on the {D}eep {I}mage {P}rior.
\newblock In {\em Conference on Computer Vision and Pattern Recognition (CVPR)}, June 2019.

\bibitem{wavelet}
H.~A. Chipman, E.~D. Kolaczyk, and R.~E. McCulloch.
\newblock {Adaptive Bayesian wavelet shrinkage}.
\newblock {\em Journal of the American Statistical Association}, 92(440):1413--1421, 1997.

\bibitem{equivariant1}
T.~Cohen and M.~Welling.
\newblock Group equivariant convolutional networks.
\newblock In {\em International Conference on Machine Learning (ICML)}, volume~48, pages 2990--2999, 2016.

\bibitem{mlp_universal3}
G.~Cybenko.
\newblock Approximation by superpositions of a sigmoidal function.
\newblock {\em Mathematics of control, signals and systems}, 2(4):303--314, 1989.

\bibitem{BM3D}
K.~Dabov, A.~Foi, V.~Katkovnik, and K.~Egiazarian.
\newblock {Image denoising by sparse 3-D transform-domain collaborative filtering}.
\newblock {\em IEEE Transactions on Image Processing}, 16(8):2080--2095, 2007.

\bibitem{tupin}
C.-A. Deledalle, L.~Denis, and F.~Tupin.
\newblock Iterative weighted maximum likelihood denoising with probabilistic patch-based weights.
\newblock {\em IEEE Transactions on Image Processing}, 18(12):2661--2672, 2009.

\bibitem{deledalle2018image}
C.-A. Deledalle, S.~Parameswaran, and T.~Q. Nguyen.
\newblock Image denoising with generalized {G}aussian mixture model patch priors.
\newblock {\em SIAM Journal on Imaging Sciences}, 11(4):2568--2609, 2018.

\bibitem{SAIST}
W.~Dong, G.~Shi, and X.~Li.
\newblock Nonlocal image restoration with bilateral variance estimation: A low-rank approach.
\newblock {\em IEEE Transactions on Image Processing}, 22(2):700--711, 2013.

\bibitem{NCSR}
W.~Dong, L.~Zhang, G.~Shi, and X.~Li.
\newblock Nonlocally centralized sparse representation for image restoration.
\newblock {\em IEEE Transactions on Image Processing}, 22(4):1620--1630, 2013.

\bibitem{nlmeans_parameters}
V.~Duval, J.-F. Aujol, and Y.~Gousseau.
\newblock A bias-variance approach for the nonlocal means.
\newblock {\em SIAM Journal on Imaging Sciences}, 4(2):760--788, 2011.

\bibitem{eckart}
C.~Eckart and G.~Young.
\newblock The approximation of one matrix by another of lower rank.
\newblock {\em Psychometrika}, 1(3):211--218, 1936.

\bibitem{tweedie}
B.~Efron.
\newblock Tweedie’s formula and selection bias.
\newblock {\em Journal of the American Statistical Association}, 106(496):1602--1614, 2011.

\bibitem{ksvd}
M.~Elad and M.~Aharon.
\newblock Image denoising via sparse and redundant representations over learned dictionaries.
\newblock {\em IEEE Transactions on Image processing}, 15(12):3736--3745, 2006.

\bibitem{ReviewElad2023}
M.~Elad, B.~Kawar, and G.~Vaksman.
\newblock Image denoising: The deep learning revolution and beyond—a survey paper.
\newblock {\em SIAM Journal on Imaging Sciences}, 16(3):1594--1654, 2023.

\bibitem{dip_rita}
R.~Fermanian, M.~Le~Pendu, and C.~Guillemot.
\newblock Regularizing the {D}eep {I}mage {P}rior with a learned denoiser for linear inverse problems.
\newblock In {\em International Workshop on Multimedia Signal Processing (MMSP)}, pages 1--6, 2021.

\bibitem{jpeg_deblocking}
A.~Foi, V.~Katkovnik, and K.~Egiazarian.
\newblock Pointwise shape-adaptive dct for high-quality deblocking of compressed color images.
\newblock In {\em European Signal Processing Conference (EUSIPCO)}, pages 1--5, 2006.

\bibitem{poisson_estimation}
A.~Foi, M.~Trimeche, V.~Katkovnik, and K.~Egiazarian.
\newblock Practical {Poissonian-Gaussian} noise modeling and fitting for single-image raw-data.
\newblock {\em IEEE Transactions on Image Processing}, 17(10):1737--1754, 2008.

\bibitem{equivariant_pointcloud2}
F.~Fuchs, D.~Worrall, V.~Fischer, and M.~Welling.
\newblock {SE(3)-Transformers}: {3D} roto-translation equivariant attention networks.
\newblock In {\em Advances in Neural Information Processing Systems (NeurIPS)}, volume~33, pages 1970--1981, 2020.

\bibitem{equivariant_translation1}
K.~Fukushima.
\newblock Neocognitron: A self-organizing neural network model for a mechanism of pattern recognition unaffected by shift in position.
\newblock {\em Biological Cybernetics}, 36(4):193--202, 1980.

\bibitem{mlp_universal2}
K.-I. Funahashi.
\newblock On the approximate realization of continuous mappings by neural networks.
\newblock {\em Neural Networks}, 2(3):183--192, 1989.

\bibitem{GemanYang02}
D.~Geman and C.~Yang.
\newblock Nonlinear image recovery with half-quadratic regularization.
\newblock {\em IEEE Transactions on Image Processing}, 4(7):932--946, 1995.

\bibitem{matching_pursuit2}
I.~Gorodnitsky and B.~Rao.
\newblock Sparse signal reconstruction from limited data using {FOCUSS}: a re-weighted minimum norm algorithm.
\newblock {\em IEEE Transactions on Signal Processing}, 45(3):600--616, 1997.

\bibitem{WNNM}
S.~Gu, L.~Zhang, W.~Zuo, and X.~Feng.
\newblock {Weighted nuclear norm minimization with application to image denoising}.
\newblock In {\em Conference on Computer Vision and Pattern Recognition (CVPR)}, pages 2862--2869, 2014.

\bibitem{equivariant_image4}
D.~K. Gupta, D.~Arya, and E.~Gavves.
\newblock Rotation equivariant siamese networks for tracking.
\newblock In {\em Conference on Computer Vision and Pattern Recognition (CVPR)}, pages 12357--12366, 2021.

\bibitem{photon_noise}
S.~W. Hasinoff.
\newblock Photon, {Poisson} noise.
\newblock {\em Computer Vision, A Reference Guide}, 4(16):1, 2014.

\bibitem{resnet}
K.~He, X.~Zhang, S.~Ren, and J.~Sun.
\newblock Deep residual learning for image recognition.
\newblock In {\em Conference on Computer Vision and Pattern Recognition (CVPR)}, pages 770--778, 2016.

\bibitem{nlridge}
S.~Herbreteau and C.~Kervrann.
\newblock Towards a unified view of unsupervised non-local methods for image denoising: the {NL-Ridge} approach.
\newblock In {\em IEEE International Conference on Image Processing (ICIP)}, pages 3376--3380, 2022.

\bibitem{lichi}
S.~Herbreteau and C.~Kervrann.
\newblock Unsupervised linear and iterative combinations of patches for image denoising.
\newblock {\em arXiv preprint arXiv:2212.00422}, 2022.

\bibitem{SHneurips23}
S.~Herbreteau, E.~Moebel, and C.~Kervrann.
\newblock Normalization-equivariant neural networks with application to image denoising.
\newblock {\em Advances in Neural Information Processing Systems (NeurIPS)}, 2023.

\bibitem{PoE}
G.~Hinton.
\newblock Products of experts.
\newblock In {\em International Conference on Artificial Neural Networks (ICANN)}, volume~1, pages 1--6, 1999.

\bibitem{rank_unconstrained}
J.-B. Hiriart-Urruty and H.~Y. Le.
\newblock From eckart and young approximation to moreau envelopes and vice versa.
\newblock {\em RAIRO-Operations Research-Recherche Op{\'e}rationnelle}, 47(3):299--310, 2013.

\bibitem{mlp_universal1}
K.~Hornik, M.~Stinchcombe, and H.~White.
\newblock Multilayer feedforward networks are universal approximators.
\newblock {\em Neural Networks}, 2(5):359--366, 1989.

\bibitem{PLR}
H.~Hu, J.~Froment, and Q.~Liu.
\newblock A note on patch-based low-rank minimization for fast image denoising.
\newblock {\em Journal of Visual Communication and Image Representation}, 50:100--110, 2018.

\bibitem{EPLL_IPOL}
S.~Hurault, T.~Ehret, and P.~Arias.
\newblock {EPLL}: an image denoising method using a {G}aussian mixture model learned on a large set of patches.
\newblock {\em {Image Processing On Line}}, 8:465--489, 2018.

\bibitem{batchnorm}
S.~Ioffe and C.~Szegedy.
\newblock Batch normalization: Accelerating deep network training by reducing internal covariate shift.
\newblock In {\em International Conference on Machine Learning (ICML)}, volume~37, pages 448--456, 2015.

\bibitem{OWF}
Q.~Jin, I.~Grama, C.~Kervrann, and Q.~Liu.
\newblock Nonlocal means and optimal weights for noise removal.
\newblock {\em SIAM Journal on Imaging Sciences}, 10(4):1878--1920, 2017.

\bibitem{rethinking}
Y.~Jo, S.~Y. Chun, and J.~Choi.
\newblock Rethinking {D}eep {I}mage {P}rior for denoising.
\newblock In {\em International Conference on Computer Vision (ICCV)}, pages 5087--5096, 2021.

\bibitem{equivariant_graph2}
N.~Keriven and G.~Peyr\'{e}.
\newblock Universal invariant and equivariant graph neural networks.
\newblock In {\em Advances in Neural Information Processing Systems (NeurIPS)}, volume~32, 2019.

\bibitem{PEWA}
C.~Kervrann.
\newblock {PEWA}: Patch-based exponentially weighted aggregation for image denoising.
\newblock In {\em Advances in Neural Information Processing Systems (NIPS)}, volume~27, 2014.

\bibitem{nlmeans_kervrann}
C.~Kervrann and J.~Boulanger.
\newblock Optimal spatial adaptation for patch-based image denoising.
\newblock {\em IEEE Transactions on Image Processing}, 15(10):2866--2878, 2006.

\bibitem{localA}
C.~Kervrann and J.~Boulanger.
\newblock Local adaptivity to variable smoothness for exemplar-based image regularization and representation.
\newblock {\em International Journal of Computer Vision}, 79(1):45--69, 2008.

\bibitem{bayesnlmeans_kervrann}
C.~Kervrann, J.~Boulanger, and P.~Coup{\'e}.
\newblock Bayesian non-local means filter, image redundancy and adaptive dictionaries for noise removal.
\newblock In {\em International Conference on Scale Space and Variational Methods in Computer Vision}, pages 520--532, 2007.

\bibitem{noise2score}
K.~Kim and J.~C. Ye.
\newblock {Noise2Score}: {Tweedie’s} approach to self-supervised image denoising without clean images.
\newblock In {\em Advances in Neural Information Processing Systems (NeurIPS)}, volume~34, pages 864--874, 2021.

\bibitem{adam}
D.~Kingma and J.~Ba.
\newblock Adam: A method for stochastic optimization.
\newblock In {\em International Conference on Learning Representations (ICLR)}, 2015.

\bibitem{N2V}
A.~Krull, T.-O. Buchholz, and F.~Jug.
\newblock {Noise2Void - Learning denoising from single noisy images}.
\newblock In {\em Conference on Computer Vision and Pattern Recognition (CVPR)}, pages 2124--2132, 2019.

\bibitem{pN2V}
A.~Krull, T.~Vicar, M.~Prakash, M.~Lalit, and F.~Jug.
\newblock Probabilistic {Noise2Void}: Unsupervised content-aware denoising.
\newblock {\em Frontiers in Computer Science}, 2:5, 2020.

\bibitem{laine}
S.~Laine, T.~Karras, J.~Lehtinen, and T.~Aila.
\newblock High-quality self-supervised deep image denoising.
\newblock In {\em Advances in Neural Information Processing Systems (NeurIPS)}, volume~32, 2019.

\bibitem{nlbayes}
M.~Lebrun, A.~Buades, and J.~M. Morel.
\newblock {A nonlocal Bayesian image denoising algorithm}.
\newblock {\em SIAM Journal on Imaging Sciences}, 6(3):1665--1688, 2013.

\bibitem{noise_clinic}
M.~Lebrun, M.~Colom, and J.-M. Morel.
\newblock {The Noise Clinic}: a blind image denoising algorithm.
\newblock {\em {Image Processing On Line}}, 5:1--54, 2015.

\bibitem{equivariant_translation2}
Y.~LeCun, B.~Boser, J.~S. Denker, D.~Henderson, R.~E. Howard, W.~Hubbard, and L.~D. Jackel.
\newblock Backpropagation applied to handwritten zip code recognition.
\newblock {\em Neural Computation}, 1(4):541--551, 1989.

\bibitem{N2K}
K.~Lee and W.-K. Jeong.
\newblock {Noise2Kernel}: Adaptive self-supervised blind denoising using a dilated convolutional kernel architecture.
\newblock {\em Sensors}, 22(11):4255, 2022.

\bibitem{noise2noise}
J.~Lehtinen, J.~Munkberg, J.~Hasselgren, S.~Laine, T.~Karras, M.~Aittala, and T.~Aila.
\newblock {N}oise2{N}oise: Learning image restoration without clean data.
\newblock In {\em International Conference on Machine Learning (ICML)}, volume~80, pages 2965--2974, 2018.

\bibitem{DIP}
V.~Lempitsky, A.~Vedaldi, and D.~Ulyanov.
\newblock Deep image prior.
\newblock In {\em Conference on Computer Vision and Pattern Recognition (CVPR)}, pages 9446--9454, 2018.

\bibitem{dead2}
A.~Levin and B.~Nadler.
\newblock Natural image denoising: Optimality and inherent bounds.
\newblock In {\em Conference on Computer Vision and Pattern Recognition (CVPR)}, pages 2833--2840, 2011.

\bibitem{dead3}
A.~Levin, B.~Nadler, F.~Durand, and W.~T. Freeman.
\newblock Patch complexity, finite pixel correlations and optimal denoising.
\newblock In {\em European Conference on Computer Vision (ECCV)}, pages 73--86, 2012.

\bibitem{swinir}
J.~Liang, J.~Cao, G.~Sun, K.~Zhang, L.~Van~Gool, and R.~Timofte.
\newblock {SwinIR}: Image restoration using {Swin Transformer}.
\newblock In {\em International Conference on Computer Vision Workshops (ICCVW)}, pages 1833--1844, 2021.

\bibitem{dset_flickr2k}
B.~Lim, S.~Son, H.~Kim, S.~Nah, and K.~M. Lee.
\newblock Enhanced deep residual networks for single image super-resolution.
\newblock In {\em Conference on Computer Vision and Pattern Recognition Workshops (CVPRW)}, pages 1132--1140, 2017.

\bibitem{score_approx}
J.~H. Lim, A.~Courville, C.~Pal, and C.-W. Huang.
\newblock {AR}-{DAE}: Towards unbiased neural entropy gradient estimation.
\newblock In {\em International Conference on Machine Learning (ICML)}, volume 119, pages 6061--6071, 2020.

\bibitem{nlrn}
D.~Liu, B.~Wen, Y.~Fan, C.~C. Loy, and T.~S. Huang.
\newblock Non-local recurrent network for image restoration.
\newblock In {\em Advances in Neural Information Processing Systems (NeurIPS)}, volume~31, 2018.

\bibitem{EPLL_unsupervised}
H.~Liu, X.~Liu, J.~Lu, and S.~Tan.
\newblock Self-supervised image prior learning with {GMM} from a single noisy image.
\newblock In {\em International Conference on Computer Vision (ICCV)}, pages 2825--2834, 2021.

\bibitem{DIP_TV}
J.~Liu, Y.~Sun, X.~Xu, and U.~S. Kamilov.
\newblock Image restoration using total variation regularized deep image prior.
\newblock In {\em IEEE International Conference on Acoustics, Speech and Signal Processing (ICASSP)}, pages 7715--7719, 2019.

\bibitem{mwcnn}
P.~Liu, H.~Zhang, K.~Zhang, L.~Lin, and W.~Zuo.
\newblock {Multi-level wavelet-CNN for image restoration}.
\newblock In {\em Conference on Computer Vision and Pattern Recognition Workshops (CVPRW)}, pages 773--782, 2018.

\bibitem{swin}
Z.~Liu, Y.~Lin, Y.~Cao, H.~Hu, Y.~Wei, Z.~Zhang, S.~Lin, and B.~Guo.
\newblock {Swin Transformer}: Hierarchical vision {Transformer} using shifted windows.
\newblock In {\em International Conference on Computer Vision (ICCV)}, pages 9992--10002, 2021.

\bibitem{convnext}
Z.~Liu, H.~Mao, C.-Y. Wu, C.~Feichtenhofer, T.~Darrell, and S.~Xie.
\newblock A {ConvNet} for the 2020s.
\newblock In {\em Conference on Computer Vision and Pattern Recognition (CVPR)}, pages 11966--11976, 2022.

\bibitem{wavelet_best2}
F.~Luisier, T.~Blu, B.~Forster, and M.~Unser.
\newblock {Which wavelet bases are the best for image denoising?}
\newblock In {\em Wavelets XI}, volume 5914, page 59140E, 2005.

\bibitem{wavelet_sure}
F.~Luisier, T.~Blu, and M.~Unser.
\newblock A new {SURE} approach to image denoising: interscale orthonormal wavelet thresholding.
\newblock {\em IEEE Transactions on Image Processing}, 16(3):593--606, 2007.

\bibitem{dset_waterloo}
K.~Ma, Z.~Duanmu, Q.~Wu, Z.~Wang, H.~Yong, H.~Li, and L.~Zhang.
\newblock Waterloo exploration database: New challenges for image quality assessment models.
\newblock {\em IEEE Transactions on Image Processing}, 26(2):1004--1016, 2017.

\bibitem{LSSC}
J.~Mairal, F.~Bach, J.~Ponce, G.~Sapiro, and A.~Zisserman.
\newblock Non-local sparse models for image restoration.
\newblock In {\em International Conference on Computer Vision (ICCV)}, pages 2272--2279, 2009.

\bibitem{MakitaloTIP2011}
M.~Makitalo and A.~Foi.
\newblock A closed-form approximation of the exact unbiased inverse of the anscombe variance-stabilizing transformation.
\newblock {\em IEEE Transactions on Image Processing}, 20(9):2697--2698, 2011.

\bibitem{matching_pursuit}
S.~Mallat and Z.~Zhang.
\newblock Matching pursuits with time-frequency dictionaries.
\newblock {\em IEEE Transactions on Signal Processing}, 41(12):3397--3415, 1993.

\bibitem{nlmeans3}
J.~V. Manj{\'o}n, J.~Carbonell-Caballero, J.~J. Lull, G.~Garc{\'\i}a-Mart{\'\i}, L.~Mart{\'\i}-Bonmat{\'\i}, and M.~Robles.
\newblock {MRI denoising using non-local means}.
\newblock {\em Medical image analysis}, 12(4):514--523, 2008.

\bibitem{red30}
X.~Mao, C.~Shen, and Y.-B. Yang.
\newblock Image restoration using very deep convolutional encoder-decoder networks with symmetric skip connections.
\newblock In {\em Advances in Neural Information Processing Systems (NIPS)}, volume~29, 2016.

\bibitem{equivariant_image}
D.~Marcos, M.~Volpi, N.~Komodakis, and D.~Tuia.
\newblock Rotation equivariant vector field networks.
\newblock In {\em International Conference on Computer Vision (ICCV)}, pages 5058--5067, 2017.

\bibitem{berkeley}
D.~Martin, C.~Fowlkes, D.~Tal, and J.~Malik.
\newblock A database of human segmented natural images and its application to evaluating segmentation algorithms and measuring ecological statistics.
\newblock In {\em International Conference on Computer Vision (ICCV)}, volume~2, pages 416--423 vol.2, 2001.

\bibitem{DIP_RED}
G.~Mataev, P.~Milanfar, and M.~Elad.
\newblock {DeepRED}: Deep image prior powered by {RED}.
\newblock In {\em International Conference on Computer Vision Workshops (ICCVW)}, 2019.

\bibitem{averaging_filters}
P.~Milanfar.
\newblock Symmetrizing smoothing filters.
\newblock {\em SIAM Journal on Imaging Sciences}, 6(1):263--284, 2013.

\bibitem{homogeneous_functions}
S.~Mohan, Z.~Kadkhodaie, E.~P. Simoncelli, and C.~Fernandez-Granda.
\newblock Robust and interpretable blind image denoising via bias-free convolutional neural networks.
\newblock In {\em International Conference on Learning Representations (ICLR)}, 2020.

\bibitem{noisier2noise}
N.~Moran, D.~Schmidt, Y.~Zhong, and P.~Coady.
\newblock {Noisier2Noise}: Learning to denoise from unpaired noisy data.
\newblock In {\em Conference on Computer Vision and Pattern Recognition (CVPR)}, pages 12061--12069, 2020.

\bibitem{iterative_regularization}
S.~Osher, M.~Burger, D.~Goldfarb, J.~Xu, and W.~Yin.
\newblock An iterative regularization method for total variation-based image restoration.
\newblock {\em Multiscale Modeling \& Simulation}, 4(2):460--489, 2005.

\bibitem{R2R}
T.~Pang, H.~Zheng, Y.~Quan, and H.~Ji.
\newblock {Recorrupted-to-Recorrupted}: Unsupervised deep learning for image denoising.
\newblock In {\em Conference on Computer Vision and Pattern Recognition (CVPR)}, pages 2043--2052, 2021.

\bibitem{pytorch}
A.~Paszke, S.~Gross, F.~Massa, A.~Lerer, J.~Bradbury, G.~Chanan, T.~Killeen, Z.~Lin, N.~Gimelshein, L.~Antiga, A.~Desmaison, A.~Kopf, E.~Yang, Z.~DeVito, M.~Raison, A.~Tejani, S.~Chilamkurthy, B.~Steiner, L.~Fang, J.~Bai, and S.~Chintala.
\newblock {PyTorch}: An imperative style, high-performance deep learning library.
\newblock In {\em Advances in Neural Information Processing Systems (NeurIPS)}, volume~32, 2019.

\bibitem{n3net}
T.~Pl\"{o}tz and S.~Roth.
\newblock Neural nearest neighbors networks.
\newblock In {\em Advances in Neural Information Processing Systems (NeurIPS)}, volume~31, 2018.

\bibitem{DND}
T.~Plötz and S.~Roth.
\newblock Benchmarking denoising algorithms with real photographs.
\newblock In {\em Conference on Computer Vision and Pattern Recognition (CVPR)}, pages 2750--2759, 2017.

\bibitem{wavelet_best}
J.~Portilla, V.~Strela, M.~Wainwright, and E.~Simoncelli.
\newblock {Image denoising using scale mixtures of Gaussians in the wavelet domain}.
\newblock {\em IEEE Transactions on Image Processing}, 12(11):1338--1351, 2003.

\bibitem{convallaria}
M.~Prakash, M.~Lalit, P.~Tomancak, A.~Krul, and F.~Jug.
\newblock Fully unsupervised probabilistic {Noise2Void}.
\newblock In {\em International Symposium on Biomedical Imaging (ISBI)}, pages 154--158, 2020.

\bibitem{S2S}
Y.~Quan, M.~Chen, T.~Pang, and H.~Ji.
\newblock {Self2Self} with dropout: Learning self-supervised denoising from single image.
\newblock In {\em Conference on Computer Vision and Pattern Recognition (CVPR)}, pages 1887--1895, 2020.

\bibitem{monte_carlo_sure}
S.~Ramani, T.~Blu, and M.~Unser.
\newblock {Monte-Carlo SURE}: A black-box optimization of regularization parameters for general denoising algorithms.
\newblock {\em IEEE Transactions on Image Processing}, 17(9):1540--1554, 2008.

\bibitem{sgd}
H.~Robbins and S.~Monro.
\newblock A stochastic approximation method.
\newblock {\em The Annals of Mathematical Statistics}, pages 400--407, 1951.

\bibitem{RED}
Y.~Romano, M.~Elad, and P.~Milanfar.
\newblock {The little engine that could: Regularization by Denoising (RED)}.
\newblock {\em SIAM Journal on Imaging Sciences}, 10(4):1804--1844, 2017.

\bibitem{unet}
O.~Ronneberger, P.~Fischer, and T.~Brox.
\newblock {U-Net}: Convolutional networks for biomedical image segmentation.
\newblock In {\em Medical Image Computing and Computer Assisted Intervention (MICCAI)}, pages 234--241, 2015.

\bibitem{mlp_inventor}
F.~Rosenblatt.
\newblock The perceptron: a probabilistic model for information storage and organization in the brain.
\newblock {\em Psychological Review}, 65(6):386, 1958.

\bibitem{FoE}
S.~Roth and M.~Black.
\newblock {Fields of Experts}: a framework for learning image priors.
\newblock In {\em Conference on Computer Vision and Pattern Recognition (CVPR)}, volume~2, pages 860--867, 2005.

\bibitem{TV}
L.~Rudin, S.~Osher, and E.~Fatemi.
\newblock Nonlinear total variation based noise removal algorithms.
\newblock {\em Physica D: Nonlinear Phenomena}, 60:259--268, 1992.

\bibitem{dset_labelme}
B.~C. Russell, A.~Torralba, K.~P. Murphy, and W.~T. Freeman.
\newblock {LabelMe}: a database and web-based tool for image annotation.
\newblock {\em International Journal of Computer Vision}, 77:157--173, 2008.

\bibitem{subdip}
A.~Sagel, A.~Roumy, and C.~Guillemot.
\newblock Sub-dip: Optimization on a subspace with deep image prior regularization and application to superresolution.
\newblock In {\em IEEE International Conference on Acoustics, Speech and Signal Processing (ICASSP)}, pages 2513--2517, 2020.

\bibitem{these}
J.~Salmon.
\newblock {\em Agrégation d’estimateurs et méthodes à patch pour le débruitage d’images numériques}.
\newblock PhD thesis, Université Paris Diderot, 2010.

\bibitem{yaroslavsky}
J.~Salmon, R.~Willett, and E.~Arias-Castro.
\newblock A two-stage denoising filter: The preprocessed {Yaroslavsky} filter.
\newblock In {\em IEEE Statistical Signal Processing Workshop (SSP)}, pages 464--467, 2012.

\bibitem{equivariant_graph3}
V.~G. Satorras, E.~Hoogeboom, and M.~Welling.
\newblock E(n) equivariant graph neural networks.
\newblock In {\em International Conference on Machine Learning (ICML)}, volume 139, pages 9323--9332, 2021.

\bibitem{pixelshuffle}
W.~Shi, J.~Caballero, F.~Huszár, J.~Totz, A.~P. Aitken, R.~Bishop, D.~Rueckert, and Z.~Wang.
\newblock Real-time single image and video super-resolution using an efficient sub-pixel convolutional neural network.
\newblock In {\em Conference on Computer Vision and Pattern Recognition (CVPR)}, pages 1874--1883, 2016.

\bibitem{sure_loss}
S.~Soltanayev and S.~Y. Chun.
\newblock Training deep learning based denoisers without ground truth data.
\newblock In {\em Advances in Neural Information Processing Systems (NeurIPS)}, volume~31, 2018.

\bibitem{dropout}
N.~Srivastava, G.~Hinton, A.~Krizhevsky, I.~Sutskever, and R.~Salakhutdinov.
\newblock Dropout: A simple way to prevent neural networks from overfitting.
\newblock {\em Journal of Machine Learning Research}, 15(56):1929--1958, 2014.

\bibitem{Starck98}
J.-L. Starck and F.~Murtagh.
\newblock Automatic noise estimation from the multiresolution support.
\newblock {\em Publications of the Astronomical Society of the Pacific}, 110(744):193, 1998.

\bibitem{Anscombe}
J.-L. Starck, F.~D. Murtagh, and A.~Bijaoui.
\newblock {\em {Image Processing and Data Analysis: the multiscale approach}}.
\newblock Cambridge University Press, 1998.

\bibitem{SURE}
C.~M. Stein.
\newblock Estimation of the mean of a multivariate normal distribution.
\newblock {\em The Annals of Statistics}, 9(6):1135--1151, 1981.

\bibitem{equivariant_pointcloud1}
N.~Thomas, T.~Smidt, S.~Kearnes, L.~Yang, L.~Li, K.~Kohlhoff, and P.~Riley.
\newblock Tensor field networks: Rotation-and translation-equivariant neural networks for {3D} point clouds.
\newblock {\em arXiv preprint arXiv:1802.08219}, 2018.

\bibitem{bilateral}
C.~Tomasi and R.~Manduchi.
\newblock Bilateral filtering for gray and color images.
\newblock In {\em International Conference on Computer Vision (ICCV)}, pages 839--846, 1998.

\bibitem{convmixer}
A.~Trockman and J.~Z. Kolter.
\newblock Patches are all you need?
\newblock {\em arXiv preprint arXiv:2201.09792}, 2022.

\bibitem{attention}
A.~Vaswani, N.~Shazeer, N.~Parmar, J.~Uszkoreit, L.~Jones, A.~N. Gomez, L.~Kaiser, and I.~Polosukhin.
\newblock Attention is all you need.
\newblock In {\em Advances in Neural Information Processing Systems (NIPS)}, volume~30, 2017.

\bibitem{equivariant_image3}
B.~S. Veeling, J.~Linmans, J.~Winkens, T.~Cohen, and M.~Welling.
\newblock Rotation equivariant {CNNs} for digital pathology.
\newblock In {\em Medical Image Computing and Computer Assisted Intervention (MICCAI)}, pages 210--218, 2018.

\bibitem{PnP}
S.~V. Venkatakrishnan, C.~A. Bouman, and B.~Wohlberg.
\newblock {Plug-and-Play} priors for model based reconstruction.
\newblock In {\em IEEE Global Conference on Signal and Information Processing}, pages 945--948, 2013.

\bibitem{normalization_morel}
Y.-Q. Wang and J.-M. Morel.
\newblock Can a single image denoising neural network handle all levels of {G}aussian noise?
\newblock {\em IEEE Signal Processing Letters}, 21(9):1150--1153, 2014.

\bibitem{CARE}
M.~Weigert, U.~Schmidt, T.~Boothe, A.~Müller, A.~Dibrov, A.~Jain, B.~Wilhelm, D.~Schmidt, C.~Broaddus, S.~Culley, M.~Rocha-Martins, F.~Segovia-Miranda, C.~Norden, R.~Henriques, M.~Zerial, M.~Solimena, J.~Rink, P.~Tomancak, L.~Royer, and E.~Myers.
\newblock Content-aware image restoration: pushing the limits of fluorescence microscopy.
\newblock {\em Nature Methods}, 15(12):1090--1097, 2018.

\bibitem{equivariant_image2}
M.~Weiler, F.~A. Hamprecht, and M.~Storath.
\newblock Learning steerable filters for rotation equivariant {CNNs}.
\newblock In {\em Conference on Computer Vision and Pattern Recognition (CVPR)}, pages 849--858, 2018.

\bibitem{dilated}
F.~Yu and V.~Koltun.
\newblock Multi-scale context aggregation by dilated convolutions.
\newblock {\em arXiv preprint arXiv:1511.07122}, 2015.

\bibitem{DCT}
G.~Yu and G.~Sapiro.
\newblock {DCT} image denoising: a simple and effective image denoising algorithm.
\newblock {\em {Image Processing On Line}}, 1:292--296, 2011.

\bibitem{PLE}
G.~Yu, G.~Sapiro, and S.~Mallat.
\newblock Solving inverse problems with piecewise linear estimators: From {G}aussian mixture models to structured sparsity.
\newblock {\em IEEE Transactions on Image Processing}, 21(5):2481--2499, 2012.

\bibitem{restormer}
S.~W. Zamir, A.~Arora, S.~Khan, M.~Hayat, F.~S. Khan, and M.~Yang.
\newblock Restormer: Efficient transformer for high-resolution image restoration.
\newblock In {\em Conference on Computer Vision and Pattern Recognition (CVPR)}, pages 5718--5729, 2022.

\bibitem{cycleisp}
S.~W. Zamir, A.~Arora, S.~Khan, M.~Hayat, F.~S. Khan, M.-H. Yang, and L.~Shao.
\newblock {CycleISP}: Real image restoration via improved data synthesis.
\newblock In {\em Conference on Computer Vision and Pattern Recognition (CVPR)}, pages 2693--2702, 2020.

\bibitem{scunet}
K.~Zhang, Y.~Li, J.~Liang, J.~Cao, Y.~Zhang, H.~Tang, D.-P. Fan, R.~Timofte, and L.~V. Gool.
\newblock {Practical blind image denoising via Swin-Conv-UNet and data synthesis}.
\newblock {\em Machine Intelligence Research}, 2023.

\bibitem{drunet}
K.~Zhang, Y.~Li, W.~Zuo, L.~Zhang, L.~Van~Gool, and R.~Timofte.
\newblock {Plug-and-Play} image restoration with deep denoiser prior.
\newblock {\em IEEE Transactions on Pattern Analysis and Machine Intelligence}, 44(10):6360--6376, 2022.

\bibitem{dncnn}
K.~Zhang, W.~Zuo, Y.~Chen, D.~Meng, and L.~Zhang.
\newblock Beyond a {G}aussian denoiser: residual learning of deep {CNN} for image denoising.
\newblock {\em IEEE Transactions on Image Processing}, 26(7):3142--3155, 2017.

\bibitem{ircnn}
K.~Zhang, W.~Zuo, S.~Gu, and L.~Zhang.
\newblock {Learning deep CNN denoiser prior for image restoration}.
\newblock In {\em Conference on Computer Vision and Pattern Recognition (CVPR)}, pages 2808--2817, 2017.

\bibitem{ffdnet}
K.~Zhang, W.~Zuo, and L.~Zhang.
\newblock {FFDNet}: Toward a fast and flexible solution for {CNN}-based image denoising.
\newblock {\em IEEE Transactions on Image Processing}, 27(9):4608--4622, 2018.

\bibitem{rnan}
Y.~Zhang, K.~Li, K.~Li, B.~Zhong, and Y.~Fu.
\newblock Residual non-local attention networks for image restoration.
\newblock In {\em International Conference on Learning Representations (ICLR)}, 2019.

\bibitem{irani}
M.~Zontak and M.~Irani.
\newblock Internal statistics of a single natural image.
\newblock In {\em Conference on Computer Vision and Pattern Recognition (CVPR)}, pages 977--984, 2011.

\bibitem{EPLL}
D.~Zoran and Y.~Weiss.
\newblock From learning models of natural image patches to whole image restoration.
\newblock In {\em International Conference on Computer Vision (ICCV)}, pages 479--486, 2011.

\end{thebibliography}

\end{document}